\documentclass[10pt,twocolumn,letterpaper]{article}

\usepackage{cvpr}              % To produce the CAMERA-READY version
\usepackage{tabularx}
\usepackage{multirow}
\usepackage{mathtools}
\usepackage{float}
\usepackage{adjustbox}

\usepackage[dvipsnames]{xcolor}

\definecolor{cvprblue}{rgb}{0.21,0.49,0.74}
\usepackage[pagebackref,breaklinks,colorlinks,citecolor=cvprblue]{hyperref}

\newcommand{\phil}[1]{{\color[rgb]{0.3,0.7,0.3}{#1}}}
\newcommand{\edward}[1]{{\color{blue}{#1}}}
\newcommand{\para}[1]{\vspace{.05in}\noindent\textbf{#1}}
\newcolumntype{C}[1]{>{\centering\arraybackslash}p{#1}}

\newcommand{\nickname}{Make-A-Shape\xspace} 
\def\ie{\emph{i.e.}}
\def\eg{\emph{e.g.}}

\def\paperID{*****} % *** Enter the Paper ID here
\def\confName{CVPR}
\def\confYear{2024}

\title{\nickname: a Ten-Million-scale 3D Shape Model}
\author{
    Ka-Hei Hui\textsuperscript{*}$^{1}$ \hspace{30pt} 
    Aditya Sanghi\textsuperscript{*}$^{2}$ \hspace{30pt} 
    Arianna Rampini$^{2}$ \hspace{30pt} 
    Kamal Rahimi Malekshan$^{2}$  \\  
    Zhengzhe Liu$^{1}$  \hspace{40pt}
    Hooman Shayani$^{2}$ \hspace{40pt} 
    Chi-Wing Fu$^{1}$ \\
    \small\textsuperscript{*}These authors contributed equally.\\
    \small$^{1}$The Chinese University of Hong Kong\hspace{40pt} 
    \small$^{2}$Autodesk AI Lab\\
    \textcolor{magenta}{\href{https://edward1997104.github.io/make-a-shape/}{https://edward1997104.github.io/make-a-shape/}}
}

\begin{document}
\twocolumn[{%
\renewcommand\twocolumn[1][]{#1}%
\maketitle
\begin{center}
    \centering
    \captionsetup{type=figure}
    \includegraphics[width=1.0\textwidth]{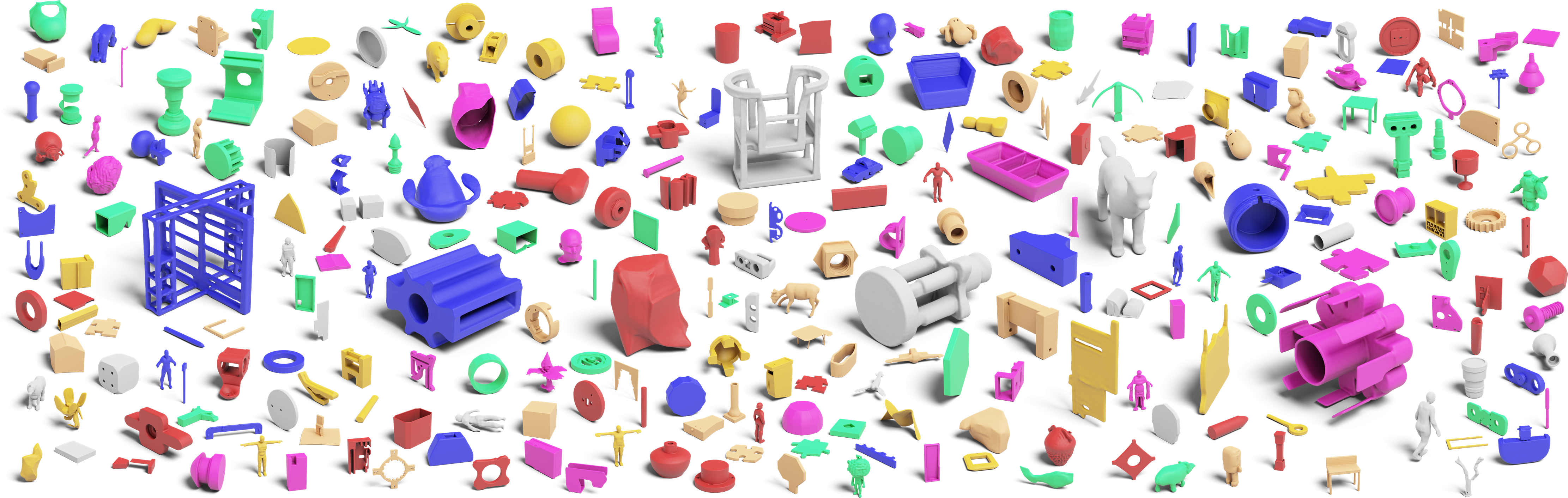}
    \captionof{figure}{ 
    \textit{\nickname} is a large 3D generative model trained on over 10 millions diverse 3D shapes.  As demonstrated above, it exhibits the capability of unconditionally generating a large variety of 3D shapes over a wide range of object categories, featuring intricate geometric details, plausible structures, nontrivial topologies, and clean surfaces.
 % \edward{TODO: We might have some discussions on this.}
 % Uncondtional generation of our method. 
 % \phil{1. Use less saturated colors. Just photoshop the image.
 % 2. I noticed that you enlarged some of the models but the enlarged models are mostly CAD models.  This may introduce a thought about whether our trained model is only or mainly for CAD.  Try to enlarge different KINDS of models.
 % 3. Also try to have more variations in the sizes of the models, rather than just mainly two sizes: large and small}
     \label{fig:teaser}
 }
\end{center}%
}]
\begin{abstract}
\vspace{-4mm}

Significant progress has been made in training large generative models for natural language and images.
Yet, the advancement of 3D generative models is %has been slower, 
hindered by their substantial resource demands for training, along with inefficient, non-compact, and less expressive representations. 
This paper introduces \nickname, a new 3D generative model designed for efficient training on a vast scale, capable of utilizing 10 millions publicly-available shapes.
Technical-wise, we first innovate a \textit{wavelet-tree representation} to compactly encode shapes by formulating the subband coefficient filtering scheme to efficiently exploit coefficient relations.
We then make the representation generatable by a diffusion model by devising the subband coefficients packing scheme to layout the representation in a low-resolution grid.
%structure.
%
Further, we derive the subband adaptive training strategy to train our model to effectively learn to generate coarse and detail wavelet coefficients.
Last, we extend our framework to be controlled by additional 
input conditions to enable it to generate shapes from assorted modalities,~\eg, single/multi-view images, point clouds, and low-resolution voxels.
 In our extensive set of experiments, we demonstrate various applications, such as unconditional generation, shape completion, and conditional generation on a wide range of modalities. Our approach not only surpasses the state of the art in delivering high-quality results but also efficiently generates shapes within a few seconds, often achieving this in just 2 seconds for most conditions.
 Our source code is available at \url{https://github.com/AutodeskAILab/Make-a-Shape}.

\end{abstract}   
 \vspace{-4mm}
\begin{figure*}[t]
	\centering
	\includegraphics[width=1.03\linewidth]{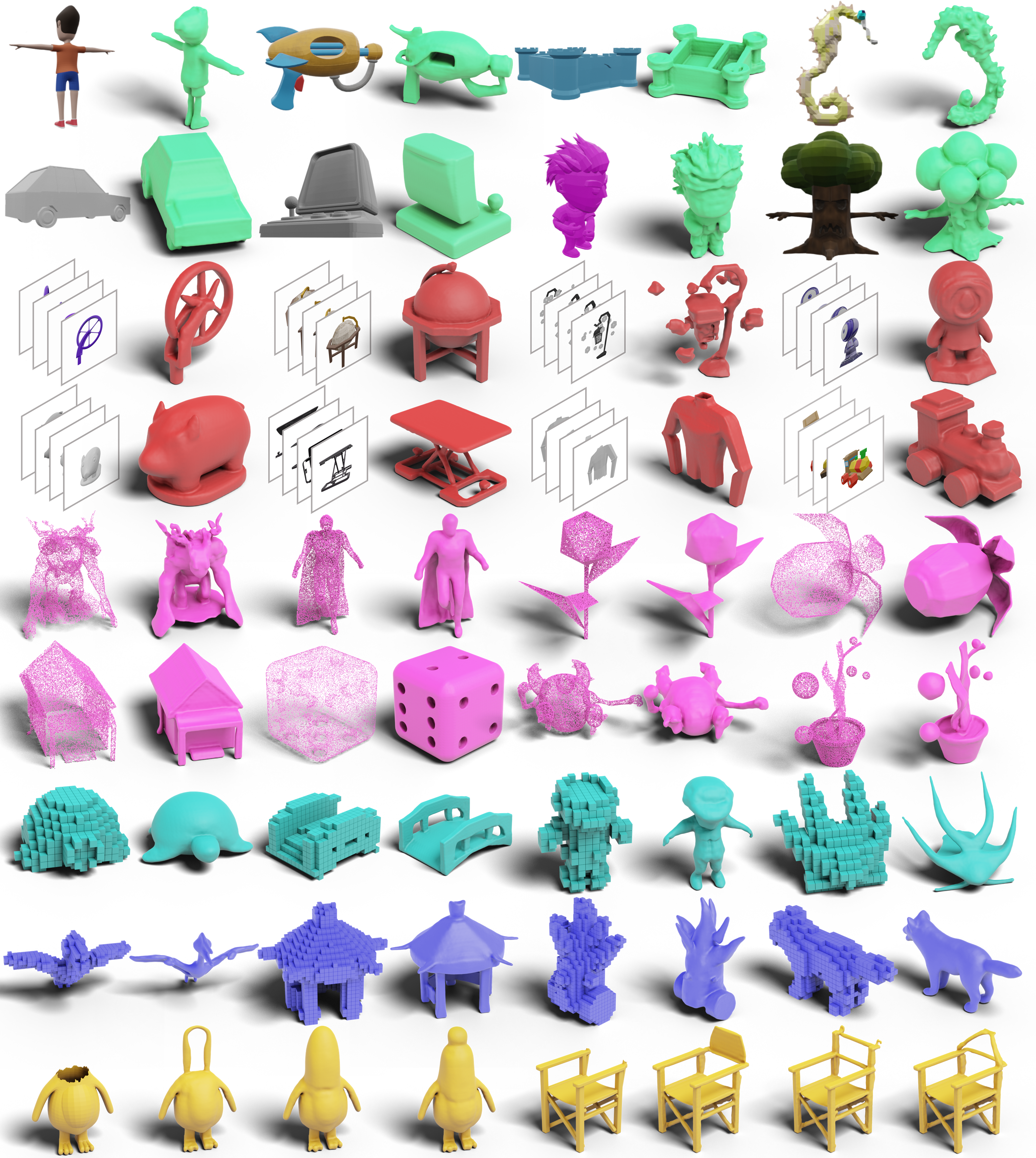}

	\caption{
\nickname is able to generate a large variety of shapes for diverse input modalities:
single-view images (rows 1 \& 2), multi-view images (rows 3 \& 4), point clouds (rows 5 \& 6), voxels (rows 7 \& 8), and incomplete inputs (last row).
The resolution of the voxels in rows 7 \& 8 are $16^3$ and $32^3$, respectively.
In the top eight rows, odd columns show the inputs whereas even columns show the generated shapes.
In the last row, columns 1 \& 4 show the partial input whereas the remaining columns show the diverse completed shapes.
 }
\label{fig:main_gallery}
\end{figure*}

% \begin{figure*}[p]
%     \centering
%     \adjustbox{height=\paperheight, keepaspectratio}{%
%         \includegraphics{Build3d_cvpr/figures/main_gallery.png}
%     }
%     \caption{Our generative model results.}
%     \label{fig:main_gallery}
% \end{figure*}
\section{Introduction}

% \phil{perhaps start the introduction by defining what large generative model is? Or talk about WHY we need large generative model?}
Large-scale generative models have increasingly become highly capable of generating realistic outputs \citep{rombach2022high, saharia2022photorealistic, ramesh2021zero, yu2022scaling}, leading to numerous commercial applications in design, marketing, e-commerce, and more.  
This success has led to the question of whether it is possible to develop a large-scale 3D generative model, which could potentially exhibit intriguing emergent properties and facilitate a variety of applications.
However, most current 3D generative models have lagged behind, being either limited in quality, focused on small 3D datasets \citep{chen2019learning, zhang2023vec, shue20233d, hui2022neural, mescheder2019occupancy, jayaraman2022solidgen, gao2022get3d, yang2019pointflow} or 
allowing a
%focusing on 
single condition~\cite{nichol2022point, jun2023shap, liu2023one, li2023instant3d, hong2023lrm, xu2023dmv3d}.

Training large generative models in 3D, compared to 2D images, faces several significant challenges. First, having an extra spatial dimension in 3D substantially increases the number of input variables that require a neural network to model, resulting far more network parameters. This is particularly evident in U-Net-based diffusion models~\citep{ho2020denoising, sohl2015deep, song2019generative}, 
%that utilize the UNET architecture~\citep{ronneberger2015u}, 
which generate memory-intensive feature maps that are often too large for GPUs to process, thus prolonging the training time~\citep{hoogeboom2023simple}. Second, scaling a generative model to 3D introduces data handling complexities not present with 2D images. Most storage and data handling for training large models takes place on cloud services such as AWS or Azure.
3D data escalates the cost of storage and time to download the data in each training iteration.
Third, there are many
ways to represent 3D shapes.
It remains unclear which one best achieves high representation quality while maintaining a good representation compactness for efficient training.

\begin{figure}[!t]
	\centering
	\includegraphics[width=1.0\columnwidth]{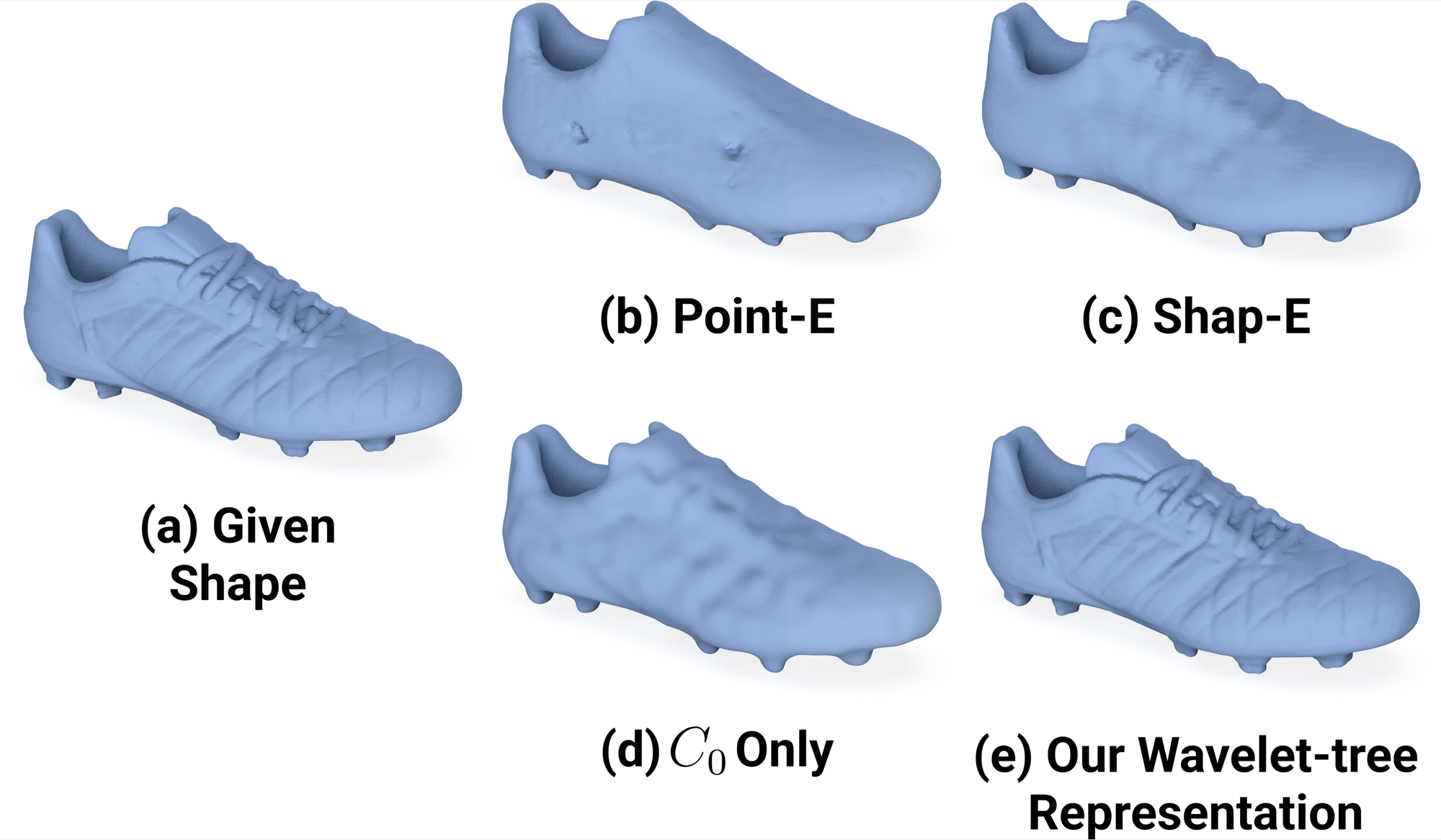}
	\caption{
        % We illustrate the reconstruction result of a given shape (a) 
        % Reconstructing a sign distance function of  shape (a) using different representations: (b) Point-E~\cite{nichol2022point}, (c) Shap-E~\cite{jun2023shap}, (d) using only coarse component $C_0$~\cite{hui2022neural}, and (e) using $C_0$ together with the detail component. 
        % Compared with others, our representation (e) can faithfully reproduce the shape with both structures and details.
        Reconstructing the SDF of a shape (a) using different methods: (b) Point-E~\cite{nichol2022point}, (c) Shap-E~\cite{jun2023shap}, (d) coarse 
        coefficients $C_0$~\cite{hui2022neural},   % Philip: note that we haven't defined what C_0 is right now
        %$C_0$~\cite{hui2022neural},
        and (e) our wavelet-tree representation. Our approach (e) can more faithfully reconstruct the
        shape's structure and details.
        % The reconstructions from Point-E, Shap-E, and $C_0$ miss some details. 
        %
        % However, our representation, enhanced with the detail component, preserves these details faithfully.
        % \phil{{\color{red}{TODO}}: (i) to change color/rendering style to make the difference or the contrast more clearly seen;
        % (ii) The text labels inside the image are too large in size;
        % (iii) $C_0$ instead of $C0$; detail component instead of detail coefficient set for consistency with the main text
        % \edward{only item (i) is left.}
        }
% }
	\label{fig:gt_recon}
\end{figure}

Recent 
large-scale generative models for 3D shapes tackle these issues through two main strategies. 
The first employs lossy input representations to
%in the modeling process.
%This approach 
reduce the number of input variables that the model must process.
However, it comes at the expense of omitting important details, thus failing to faithfully capture the shape.
Key examples of this strategy are Point-E~\citep{nichol2022point}, which utilizes point clouds as input, and Shap-E~\cite{jun2023shap}, which uses latent vectors.
Their trade-off is evident in Figure~\ref{fig:gt_recon} and Table~\ref{tab:rep_compare}, where a significant loss of detail is often observed when reconstructing the ground-truth signed distance function (SDF).
The second strategy 
%in large-scale generative modeling of 3D shapes 
employs multi-view images
%, exclusively 
to represent the geometry~\citep{liu2023one, li2023instant3d, hong2023lrm, xu2023dmv3d}. 
In this approach, a generative network utilizes differentiable rendering to produce images of the generated shape for comparing
with ground-truth multi-view images to facilitate learning.
These methods generally require extensive training time, as they use differentiable rendering for loss calculation, which can be slow and may not capture full geometry in one training example.
Our framework, on average, processes 2x to 6x more training shapes in one day than these methods, despite utilizing a less powerful GPU (A10G vs. A100), as detailed in Table~\ref{tab:inf_compare_baseline}.

These issues arise due to the lack of 
%using 
a suitable 3D representation that is expressive, compact, and efficient to learn.
In this work, we introduce a new 3D representation, the \textit{wavelet-tree representation}, designed for encoding 3D shapes for large-scale model training.
This representation employs a wavelet decomposition on a high-resolution SDF grid to yield a coarse coefficient subband and multiple multiscale detail coefficient subbands.
Beyond~\cite{hui2022neural}, which discards all detail subbands for efficient generation, we design a family of techniques to enable large model training, considering both coarse and detail coefficients:
(i) {\em subband coefficients filtering\/} to identify and retain 
information-rich detail coefficients in the detail subbands, such that our representation can compactly include more shape details;
(ii) {\em subband coefficients packing\/} to rearrange the wavelet-tree representation in a low-resolution spatial grid, such that the re-arranged representation can become diffusible,~\ie, generatable by a diffusion model; and
(iii) {\em subband adaptive training strategy} to enable efficient model training on both coarse and detail coefficients, such that the training can attend to the overall shape and also the important but sparse shape details.
Besides, we formulate various conditioning mechanisms to accommodate flexibly input conditions, such as point clouds, voxels, and images. 
Hence, our new representation, while being compact, can faithfully retain most shape information and facilitate effective training of a large generative model on over millions of 3D shapes.

\begin{table}[t]
\caption{
% Comparison of different 3D representations. The ``IOU'' column indicates the average intersection over union (IOU) for reconstructions of 1030 shapes on the GSO dataset.
% %
% The ``Input Variable'' column shows the number of floating points adopted for each representation.
% %
% The ``Extra training'' and ``Process Time'' columns indicate whether an additional network is needed for reconstructing a 3D shape and the time taken to convert a 3D shape (an SDF volume) to a corresponding representation, respectively.
Comparing different 3D representations on the GSO dataset~\cite{downs2022google} in terms of Intersection Over Union (IOU) and processing time.
``Extra Network'' denotes the need for training multiple networks to obtain SDF;
``Process Time'' refers to the time required to convert from one representation to SDF; and
``Input Variable'' reports the number of floating-point numbers adopted in each representation. It is noteworthy that our representation has a similar parameter count as Shap-E~\cite{jun2023shap}, yet it does not need an extra network and achieves faster conversion.
}
\label{tab:rep_compare}
\centering
{\small
\resizebox{1.0\linewidth}{!}{
\begin{tabular}{c|c|c|c|c}
\toprule
Representation            & IOU & Input Variables & Extra Network & Process Time \\
\midrule
Ground-truth SDF ($256^3$) & 1.0                       & 16777216         & No                        & $-$ \\ 
Point-E~\cite{nichol2022point}                   & 0.8642                       & 12288           & Yes                       & $\sim$1 second  \\
Shap-E~~\cite{jun2023shap}                    & 0.8576                       & 1048576         & Yes                       & $\sim$5 minutes \\
Coarse Component~\cite{hui2022neural}     & 0.9531                       & 97336           & No                        & $\sim$1 second  \\
Wavelet tree (ours) & 0.9956                       & 1129528         & No                        & $\sim$1 second \\ 

\bottomrule
\end{tabular}
}
}
\end{table}
%This proposed 
With the above technical contributions, we can generate a representation that is notably \textit{expressive}, capable of encoding a shape with minimal loss; for example, a $256^3$ grid can be bijectively encoded into the wavelet-tree representation in around one second, yet with an IoU of 99.56\%.
Simultaneously, our representation is \textit{compact}, characterized by a low number of input variables. This is almost akin to lossy representations like latent vectors~\cite{jun2023shap}, yet it achieves this without necessitating additional training of an autoencoder, while having higher quality, as shown in Table~\ref{tab:rep_compare}.
Last, our representation is \textit{efficient}, enabling efficient streaming and training.
For instance, streaming and loading a sophisticatedly compressed $256^3$ SDF grid takes 266 milliseconds, while our representation requires only 184 milliseconds for the same process. 
The 44.5\% reduction in data loading time is crucial for large-scale model training.

\begin{table}[t]
\centering
{\fontsize{8}{9.6}\selectfont 
\caption{Efficiency comparison with 
%other 
state-of-the-art methods. The results for rows 4-6 are sourced from the concurrent works, DMV3D~\cite{xu2023dmv3d}, Instant3D~\cite{li2023instant3d}, and LRM~\cite{hong2023lrm}, respectively.
For single-view, we present inference time for both 10 and 100 iterations (iter.), with the latter being the optimal hyperparameter for quality, as per our ablation study. For multi-view, 10 iterations is identified as the optimal.
For the training time, since different methods use different number of GPUs, we compare their training speed by the number of training shapes that it can process in one day divided by the number of GPUs used.
Note that training time is not available for Point-E~\cite{nichol2022point} and Shap-E~\cite{jun2023shap}.
}
\label{tab:inf_compare_baseline}

\resizebox{1.0\linewidth}{!}{
\begin{tabular}{c@{\hspace{1mm}}|@{\hspace{1mm}}c@{\hspace{1mm}}|@{\hspace{1mm}}c}
\toprule
Method  & Inference time & \# Training shapes in 1 day / GPU\\
\midrule
Point-E \citep{nichol2022point}  & $\sim$ 31 sec & $-$ \\
Shape-E \citep{jun2023shap}  &  $\sim$ 6 sec & $-$ \\
One-2-3-45 \citep{liu2023one} & $\sim$ 45 sec & $\sim$ 50k (A10G) \\
DMV3D \citep{xu2023dmv3d} & $\sim$ 30 sec & $\sim$ 110k (A100)\\
Instant3D \citep{li2023instant3d} & $\sim$ 20 sec & $\sim$ 98k (A100)\\
LRM \citep{hong2023lrm} & $\sim$ 5 sec & $\sim$ 74k (A100) \\
% Zero123 \citep{liu2023zero}  & A100 & 15 min \\
% DreamGuassian   \citep{tang2023dreamgaussian} & V100 & 2 min \\
\midrule
Ours  (single-view 10 iter.)  & $\sim$ 2 sec &\\
Ours (single-view 100 iter.)  & $\sim$ 8 sec & \multirow{-2}{*}{$\sim$ 290k (A10G)}\\
\midrule
Ours (multi-view 10 iter.) & $\sim$ 2 sec & $\sim$ 250k (A10G) \\
\bottomrule
\end{tabular}
}
}
\end{table}

% \begin{table*}[t]
% \caption{Inference Time Comparison}
% \label{tab:inf_compare_baseline}
% \centering
% \small
% \begin{tabular}{c|c|c|c|c|c}
% \toprule
%  & Point-E & Shape-E & One-2-3-45 & \multicolumn{2}{c}{Ours} \\
%  & \citep{nichol2022point} & \citep{jun2023shap} & \citep{liu2023one} & SV & MV \\
% \midrule
% GPU & & & & \multicolumn{2}{c}{} \\
% \midrule
% Inference Time & 78 sec & 27 sec & 45 sec & 7 sec & 1 sec \\
% \bottomrule
% \end{tabular}

% \end{table*}

Overall, our generative model can be trained effectively, enabling also fast inference and taking just \textit{few seconds} to generate high-quality shapes, as compared to existing methods reported in Table~\ref{tab:inf_compare_baseline}.
We name our proposed generation framework \textit{\nickname}. This framework facilitates the training of an unconditional generative model and 
%various
extended models under different input conditions on an extensive dataset comprising 10 million 3D shapes over a wide range of object categories. It successfully produces a range of plausible shapes, as illustrated in Figures~\ref{fig:teaser} and~\ref{fig:main_gallery}.

\begin{figure*}[!t]
	\centering
	\includegraphics[width=1.0\linewidth]{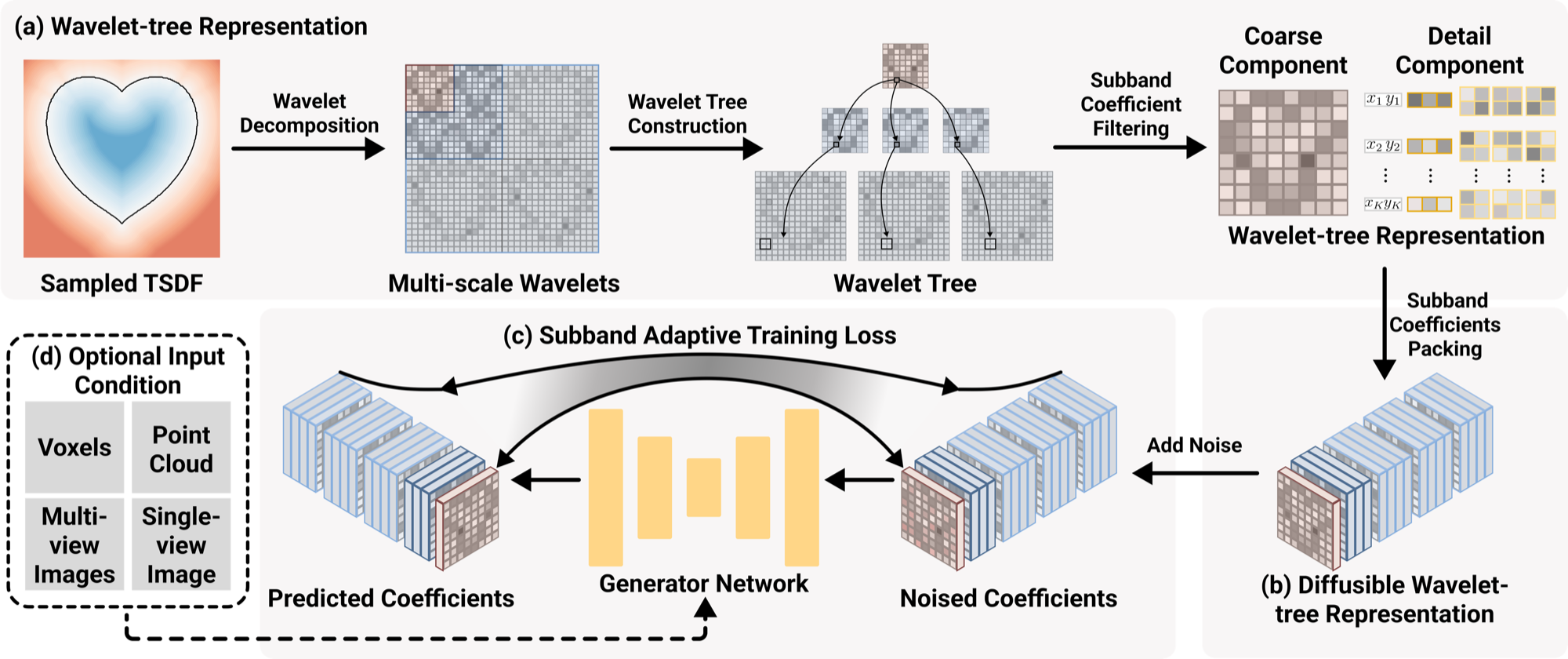}
	\caption{
 % \edward{TODO: update the colors of TSDFs.}
 Overview of our generative approach. 
 %The overview of our 3D generation learning framework. 
 %
 (a) A shape is first encoded into a truncated signed distance field (TSDF), then decomposed into multi-scale wavelet coefficients in a wavelet-tree structure.
 %
 % Using the parent-child relation between the coefficients, we can then construct a wavelet tree structure (with $D_2$ omitted).
 %
 We design the {\em subband coefficient filtering\/} procedure to exploit the relations among coefficients and extract information-rich coefficients to build our wavelet-tree representation.
 (b) We propose the {\em subband coefficient packing\/} scheme to rearrange our wavelet-tree representation into a regular grid structure of manageable spatial resolution, so that we can adopt a denoising diffusion model to effectively generate the representation.
 %from noise.
 %(b) We then pack different subband coefficients in our wavelet-tree representation into a regular grid structure with a small spatial resolution, which can be effectively generated or {\em diffusible} by a denosing diffusion probabilistic model (DDPM).
 % (b) To convert our representation into a diffusible format for a generator network, we first unpack the detail coefficient set into a dense coefficient volume, and further enhance the efficiency by concatenating corresponding coefficients to reduce input resolution.
 %
 (c) Further, we formulate the {\em subband adaptive training\/} strategy to effectively balance the shape information in different subbands and address the detail coefficient sparsity.
 Hence, we can efficiently train our model on millions of 3D shapes.
 %(c) We also propose a subband adaptive training strategy that can  effectively balance shape information in different components and address the sparsity in the detail component.
 % to train a denoising diffusion probabilistic model (DDPM) for generating our converted representation.
 %
 (d) Our framework can be extended to condition on various modalities.
 %(d) Our proposed framework can be easily extended for taking various forms of additional conditions via an extra encoder and various conditioning mechanisms.
% \phil{{\color{red}{TODO}}: please make the boxes for coarse and detail components on top-right smaller such that other elements on top row less tight, e.g., the wavelet tree part is too tight}
 }
 \label{fig:overview}
\end{figure*}
\section{Related Work}

\noindent \textbf{Neural Shape Representations.}  
In recent years, there has been significant research in integrating deep learning with 
%different 
various 3D 
%shape 
representations. Initial works such as~\citep{wu20153d, maturana2015voxnet} focused on volumetric representations for different tasks through the use of 3D convolutional networks. 
Multiview CNNs were also explored~\cite{su2015multi, qi2016volumetric}, by means of
%focus on 
first rendering 3D shapes into multiple images, upon which 2D CNNs 
%convolutional networks 
are applied for use in downstream 
%discriminative 
applications.

Subsequently, deep learning methods for point clouds were introduced
%initially explored 
in PointNet~\citep{qi2017pointnet}, and later, additional inductive biases such as convolution were adopted, as seen in~\citep{qi2017pointnet++, wang2019dynamic}. Neural networks can also be employed to represent 3D shapes by predicting either signed distance functions (SDFs) or occupancy fields. These representations, typically known as neural implicit representations, have been a popular subject of exploration~\citep{park2019deepsdf, mescheder2019occupancy, chen2019learning}. Other explicit representations such as meshes have been explored in works such as~\citep{hanocka2019meshcnn, masci2015geodesic, verma2018feastnet, nash2020polygen}. Another prevalent 3D representation, boundary representation (BREP), has only been examined recently in 
%various 
studies such as~\citep{jayaraman2021uv, lambourne2021brepnet, wu2021deepcad, jayaraman2022solidgen} for 
%numerous 
discriminative and generative applications.

Recently, some works~\cite{hui2022neural, liu2023exim} have investigated the use of wavelets to decompose an SDF signal into multi-scale wavelet coefficients. These methods, however, filter out high-frequency details to enhance learning efficiency, albeit at the expense of shape fidelity.
In this work, we introduce a novel representation known as the wavelet-tree representation. We consider both coarse and information-rich detail coefficients to compactly yet nearly losslessly encode 3D shapes.
Enabled by various techniques that we shall introduce, our representation enables high-quality shape generation, while remaining compact for scalability across a large 3D data comprising over millions of shapes.

\vspace*{2mm}
\noindent \textbf{3D Generative Models.}  
The initial efforts in the field of 3D generative models primarily concentrated on Generative Adversarial Networks (GANs) \cite{goodfellow2014generative, wu2016learning}. Subsequently, autoencoders are trained and GANs are then utilized to process the latent spaces of these autoencoders, enabling generative models on representations such as point clouds~\cite{achlioptas2018learning} and implicit representations~\cite{chen2019learning, ibing20213d, zheng2022sdf}. More recent studies~\cite{chan2022efficient, schwarz2022voxgraf, gao2022get3d} incorporated GANs with differentiable rendering, where multiple rendered views are employed as the loss signal.
There has also been a focus on normalizing flows~\cite{yang2019pointflow, klokov2020discrete, sanghi2022clip} and Variational Autoencoder (VAE)-based generative models~\cite{mo2019structurenet}. Autoregressive models have gained popularity in 3D generative modeling and have been extensively explored~\cite{cheng2022autoregressive, nash2020polygen, sun2020pointgrow, mittal2022autosdf, yan2022shapeformer, zhang20223dilg, sanghi2023clip}.

With recent advances in diffusion model for high-quality image generation, there has also been immense interest in diffusion models for 3D context. Most approaches first train a Vector-Quantized-VAE (VQ-VAE) on a 3D representation such as triplane~\cite{shue20233d, chou2023diffusion, peng2020convolutional}, implicit form~\cite{zhang20233dshape2vecset, li2023diffusion, cheng2023sdfusion} and point cloud~\cite{jun2023shap, zeng2022lion}, before employing the diffusion model to the latent space.
Direct training on a 3D representation has been less explored. Some recent studies focus on point clouds~\cite{nichol2022point, zhou20213d, luo2021diffusion}, voxels~\cite{zheng2023locally}, and neural wavelet coefficients~\cite{hui2022neural, liu2023exim}. 
Our work employs a diffusion model directly on the 3D representation, thereby avoiding information loss associated with the VQ-VAE.
Besides formulating our wavelet-tree representation, we propose a scheme to convert it into a format that can be diffusible, or effectively generatable by a diffusion model.
Our approach demonstrates great efficiency at high resolutions compared to~\cite{zheng2023locally}, produces cleaner manifold surfaces than~\cite{nichol2022point, zhou20213d, luo2021diffusion} and captures far more details than~\cite{hui2022neural}.

\noindent \textbf{Conditional 3D Models.}  
Existing conditional models in 3D can be categorized in two groups. The first group leverages large 2D conditional image generative models, such as Stable Diffusion~\cite{rombach2022high} or Imagen~\cite{saharia2022photorealistic}, to optimize a 3D scene or object. 
%This process creates a 3D shape that is converted to images using a differentiable renderer. These images are then either compared to 
These methods create 3D shapes and convert them to images using a differentiable renderer, such that the images can be either compared to 
multiple images or aligned with the distribution of a large text-to-3D generative model. The initial exploration in this area was centered around text-to-3D, as seen in~\cite{jain2022zero, michel2022text2mesh, poole2022dreamfusion}. This approach was later expanded to include images~\cite{deng2023nerdi, melas2023realfusion, xu2022neurallift} and multi-view images~\cite{liu2023zero, deitke2023objaverse, qian2023magic123, shi2023mvdream}. Recent methods have also incorporated additional conditions such as sketches~\cite{mikaeili2023sked}.
%as normals \cite{long2023wonder3d}, depth maps \cite{chung2023luciddreamer}, and 
%
Overall, this approach unavoidably requires an expensive optimization, which limits practical applications.

The second group of methods focuses on training a conditional generative model with data that is either paired with a condition or used in a zero-shot manner. Paired conditional generative models explore various conditions such as point cloud~\cite{zhang20223dilg, zhang2023vec}, image~\cite{zhang20223dilg, nichol2022point, jun2023shap, zhang2023vec}, low-resolution voxels~ \cite{chen2021decor, chen2023shaddr}, sketches~\cite{lun20173d, guillard2021sketch2mesh, gao2022sketchsampler, kong2022diffusion} and text~\cite{nichol2022point, jun2023shap}. More recently, zero-shot methods have gained popularity, with a focus on text~\cite{sanghi2022clip, sanghi2023clip, liu2022iss, xu2023dream3d} and sketches~\cite{sanghi2023sketch}.
In this work, our primary focus is on training a large, paired conditional generative model. This model offers fast generation, as it eliminates the need for scene optimization. Our approach also facilitates the easy incorporation of assorted conditions,~\eg, point clouds, low-resolution voxels, and images. Besides, it enables both unconditional applications and zero-shot tasks like shape completion.

\section{Overview}

Figure~\ref{fig:overview} provides an overview of our shape-generative framework, designed to create a large-scale 3D generative model capable of efficient training on millions of 3D shapes. The complexity of 3D data makes efficient training at this scale extremely challenging, particularly when considering the need to optimize both the quality of shape generation and the speed of training. Our approach comprises four main components, detailed in Sections~\ref{sec:wavelet_tree} to~\ref{sec:conditional_generation}.

(i) {\em Wavelet-tree representation.}
We first formulate a compact, efficient and expressive 3D representation to support large-scale shape training.
Importantly, we first encode each shape into a high-resolution truncated signed distance field (TSDF) and decompose the TSDF into multiscale wavelet coefficients.
We design a subband coefficient filtering procedure that exploits the relationships among coefficients, allowing us to retain information-rich wavelet components (both coarse and detail) in our wavelet-tree representation, enabling a faithful yet compact representation of the 3D shape for efficient storage and streaming.

(ii) {\em Diffusible Wavelet-tree Representation.}
Next, we transform the wavelet-tree representation into a format that is more compatible with diffusion models. Though our representation achieves compact shape encoding, its irregular format hinders  effective shape learning and generation.
This motivates us to design the subband coefficient packing scheme to rearrange the coefficients into a regular grid of a manageable spatial resolution for shape generation.

(iii) {\em Subband Adaptive Training Strategy.}
Further, we exploit methods to train the model on the diffusible wavelet-tree representation.
In general, shape information varies across subbands and scales, with detail coefficients being highly sparse yet rich in the shape details.
Hence, training with a standard uniform Mean Squared Error (MSE) loss 
%in training our generative model 
might lead to model collapse or inefficient learning of the details. To address this, we introduce the {\em subband adaptive training\/} strategy, which selectively focuses on coefficients in various subbands. This approach allows for an effective balance of shape information over coarse to fine subbands during the training and encourages the model to learn both the structural and detailed aspects of shapes.

(iv) {\em Extension for Conditional Generation.}
Finally, beyond unconditional generation, we extend our method 
%which captures detailed shape information, 
to support conditional generation of shapes, following conditions such as single-/multi-view images, voxels, and point clouds.
%, moving beyond just unconditional generation.
In essence, we encode the specified conditions into latent vectors and then collectively employ multiple mechanisms to inject these vectors into our generation network.

%\section{Efficient 3D Representation}
\section{Wavelet-tree Representation}
\label{sec:wavelet_tree}
\begin{figure}[t]
	\centering
	\includegraphics[width=0.8\columnwidth]{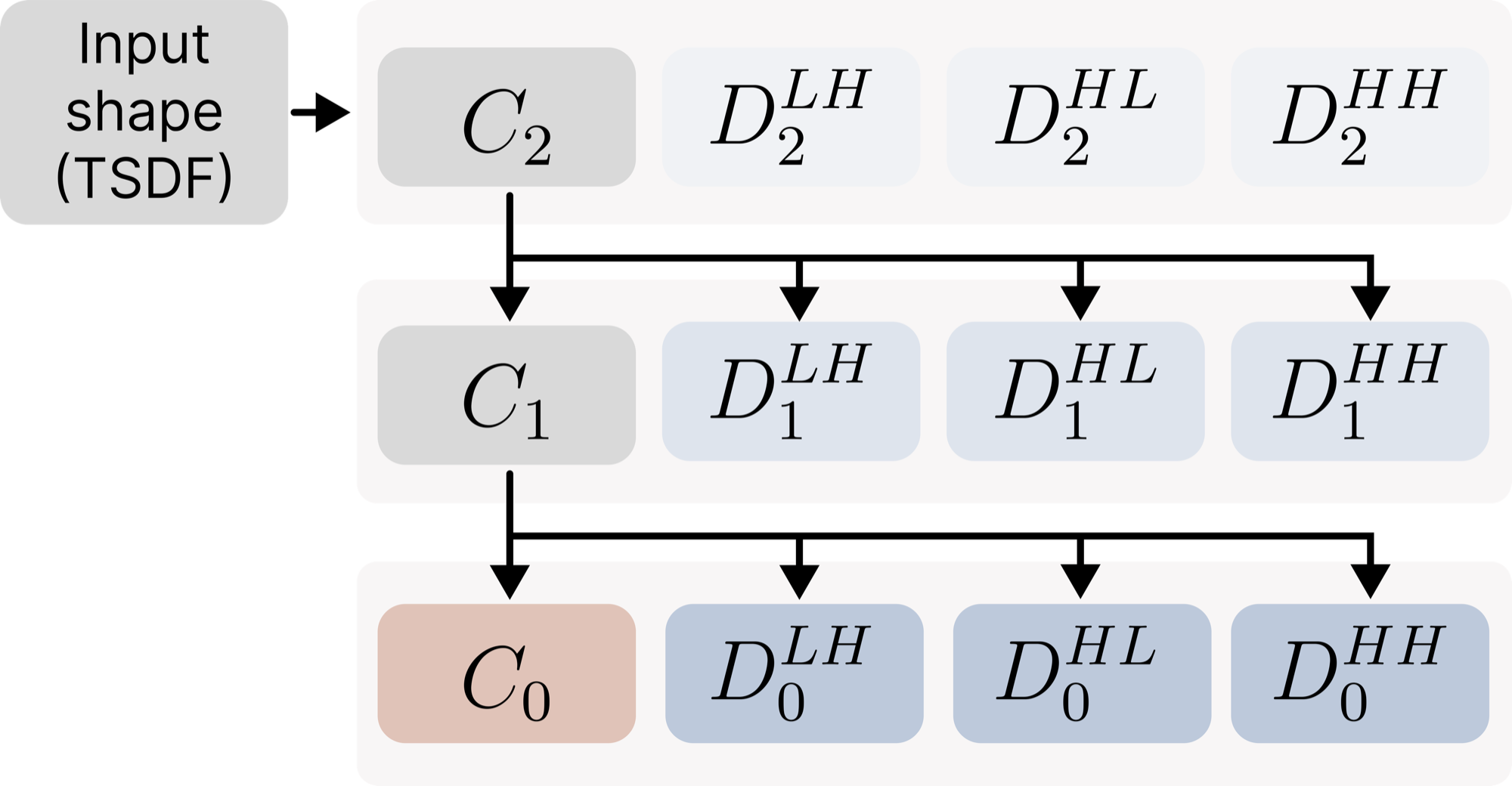}
	\caption{
Wavelet decomposition of the input shape, represented as a TSDF, recursively into coarse coefficients $C_i$ and detail coefficients $\{ D_i^{LH}, D_i^{HL}, D_i^{HH} \}$.
Note that in the 3D case, there will be seven subbands of detail coefficients in each decomposition.
%
%Wavelet decomposition procedure. Decomposing the input shape represented as a TSDF into multiscale wavelet coefficients,~\ie, coarse coefficients $C_i$ and detail coefficients $\{ D_i^{LH}, D_i^{HL}, D_i^{HH} \}$.
%Using $C_0$ and all detail coefficients, we can losslessly reconstruct the TSDF via inverse wavelet transforms.
%$\{ C_2, D_2^LH, ... \}$, then $C_2$ into $\{ C_0, D_0^LH, ... \}$ 
%, $D_1$, and $D_2$) as a lossless representation of the original TSDF.
}
\label{fig:wavelet_process}
\end{figure}

\begin{figure}[t]
	\centering
	\includegraphics[width=0.8\columnwidth]{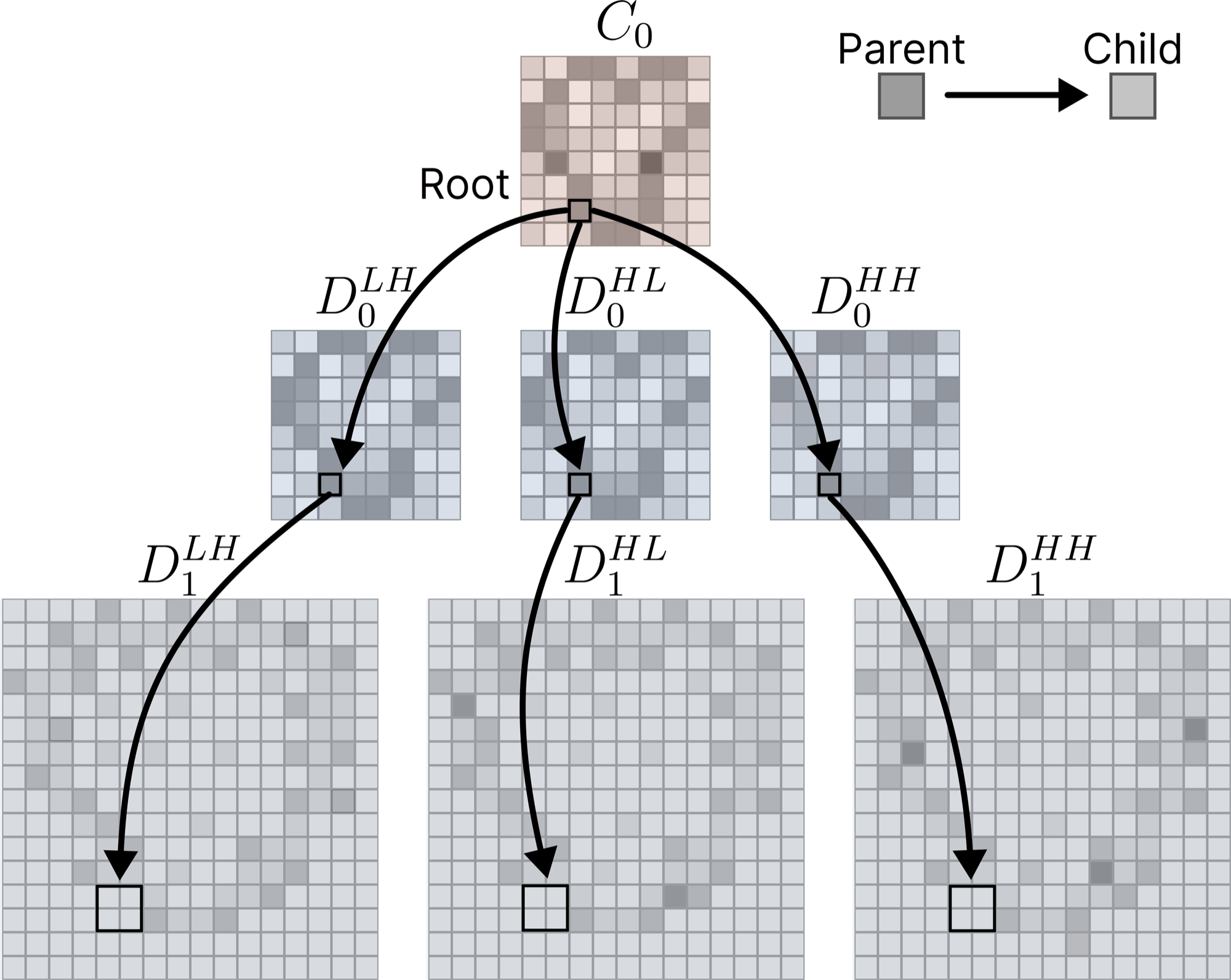}
	\caption{
Overview of Parent-child relation.
A wavelet tree is formed for each coefficient in $C_0$ as the root, with
a coarser-level coefficient as parent and the finer-level coefficients as children.
%After the multiscale wavelet decomposition, 
%through parent-child relations between 
% \phil{1. put the ``root'' text label on the left of the small black sqaure instead;
% 2. may add a legend without border (at top right side of the figure) to show a short arrow with text labels ``parent'' and ``child''}
%Wavelet tree construction of 
%
%Wavelet tree construction of multiscale wavelet coefficients, where the arrows indicate the parent-child relation between different levels of wavelet coefficients.
% \edward{Updating the font size and add text label ``root'' on the left hand side outside $C_0$.}
}
\label{fig:wavelet_tree}
\end{figure}

To build our representation, we transform a 3D shape into a truncated signed distance function (TSDF) with a resolution of $256^3$.
We decompose the TSDF using a wavelet transform\footnote{Following the approach in \cite{hui2022neural}, we use biorthogonal wavelets with 6 and 8 moments.} into a coarse coefficient $C_0$ and a set of three detail coefficients $\{D_0, D_1, D_2\}$. The process of obtaining these coefficients involves first transforming TSDF into $C_2$ and its associated detail coefficients $D_2 = { D_2^{LH}, D_2^{HL}, D_2^{HH} }$. Then, we decompose $C_2$ into $C_1$ and $D_1 = { D_1^{LH}, D_1^{HL}, D_1^{HH} }$, and subsequently, $C_1$ is decomposed into $C_0$ and $D_0 = { D_0^{LH}, D_0^{HL}, D_0^{HH} }$. This process is depicted in Figure~\ref{fig:wavelet_process}.
For simplicity, we present our method using 2D illustrations, yet the actual computation is performed in 3D with seven {\em subband volumes\/} (instead of three {\em subband images\/}, in the 2D case) of detail coefficients in each decomposition.
It is important to note that the detail coefficients contain high-frequency information. Furthermore, this representation is lossless and can be bijectively converted to a TSDF through inverse wavelet transforms.

\vspace*{-3mm}
\paragraph{Wavelet-tree and Coefficient relation.}
Building upon the neural wavelet representation as proposed in \citep{hui2022neural}, we propose to exploit the relationships between wavelet coefficients for our 3D representation. Generally, each coarse coefficient in $C_0$, referred to as a {\em parent\/}, and its associated detail coefficients in $D_0$, known as {\em children\/}, reconstruct the corresponding coefficients in $C_1$ through an inverse wavelet transform. This {\em parent-child relation\/}  relationship extends between $D_0$ and $D_1$, and so forth, as shown by the arrows leading from $D_0$ to $D_1$ in Figure~\ref{fig:wavelet_tree}. Additionally, coefficients sharing the same parent are termed {\em siblings\/}. By aggregating all descendants of a coefficient in $C_0$, we can construct a  {\em wavelet coefficient tree\/} or simply a wavelet tree, with a coefficient in $C_0$ serving as its root. This concept is further illustrated in Figure~\ref{fig:wavelet_tree}.

\vspace*{-3mm}
\paragraph{Observations.}
We have identified four notable observations regarding the wavelet coefficients:
\begin{enumerate}[label=(\roman*)]
%\begin{itemize}
%
\item
%\item[(i)]
If a coefficient's magnitude is smaller than a threshold (say, 1/32 of the largest coefficient in a subband), its children will likely have small magnitudes.
Small magnitude means low contribution to the shape, so these coefficients have little impact on the shape.
We empirically studied this observation in the $D_0$ subbands of 1,000 random shapes and found that more than $96.1 \%$ of the coefficients satisfy this hypothesis.
\item
%\item[(ii)]
The values of sibling coefficients are positively correlated.
We evaluated the correlation coefficients between all pairs of sibling coefficients in 1,000 random shapes and found a positive correlation value of 0.35.
\item
%\item[(iii)]
Coefficients in $C_0$ are mostly non-zeros, with a mean magnitude of $2.2$, while the mean magnitude of detail coefficients in $D_0$ are much closer to zero,
implying that $C_0$ contains most of the shape information.
%in the original shape.
%
\item
%\item[(iv)]
Most coefficients in $D_2$ are insignificant.
By empirically setting them to zeros in inverse wavelet transforms, we can reconstruct the TSDFs faithfully for 1,000 random shapes with $99.64\%$ IoU accuracy.
%
%\end{itemize}
\end{enumerate}

\vspace*{-3mm}
%\paragraph{Our Wavelet-tree representation.}

\begin{figure}[t]
	\centering
	\includegraphics[width=1.0\columnwidth]{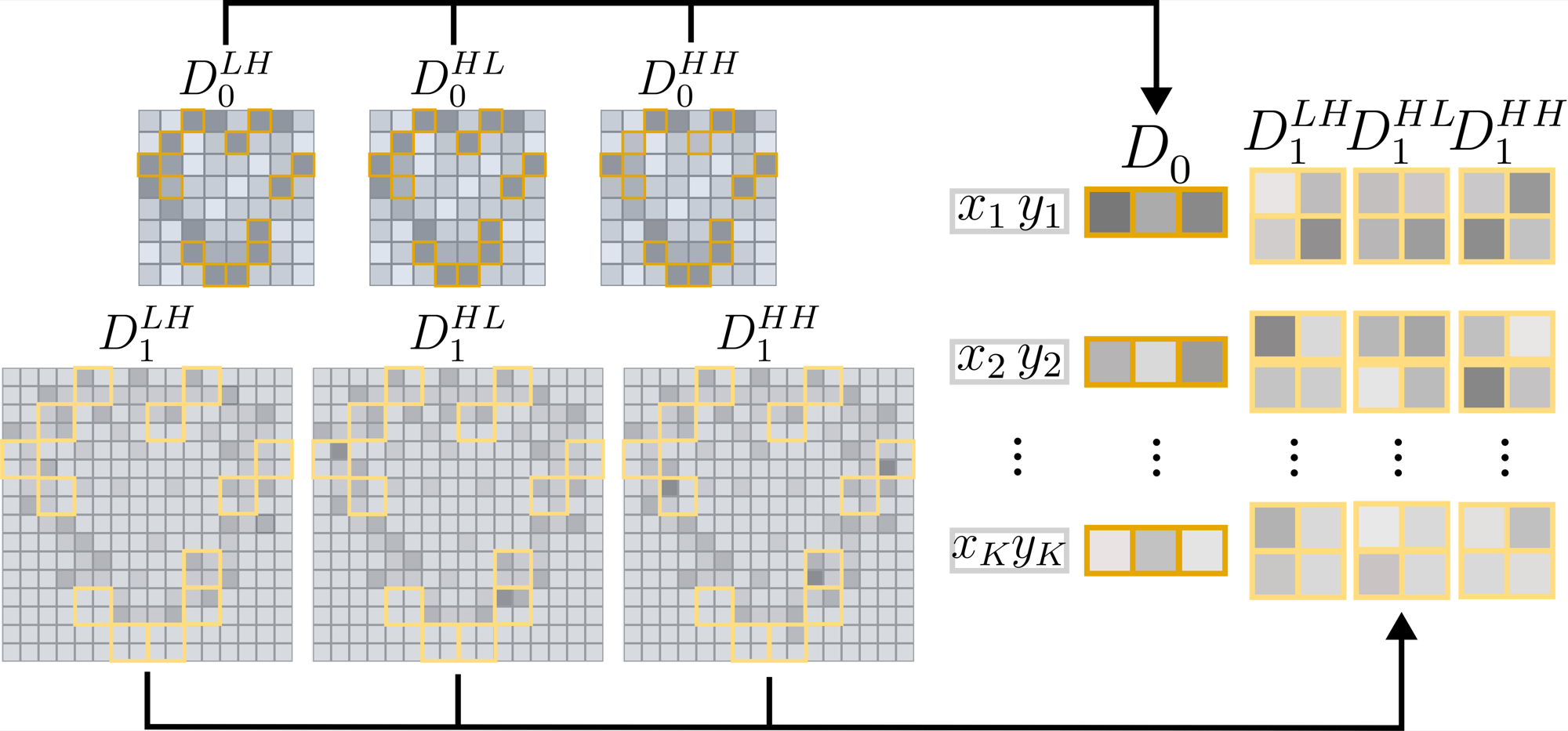}
	\caption{
 %Constructing the 
The detail component part of our representation.
We extract and pack informative coefficients from $D_0$ and $D_1$, indicated in yellow boxes, along with their spatial locations to form our representation's detail component.
% We extract informative coefficients (marked in yellow boxes) in $D_0$ and $D_1$ (left), then pack their spatial locations and associated coefficient values to form the detail component (right) in our representation.
 % \edward{I have another version with a horizontal layout, but it still looks too small.}
% After the multiscale wavelet decomposition, we can form a wavelet tree with each coefficient in $C_0$ as its root through parent-child relations between a coarser-level coefficient (parent) and finer-level coefficients (children).
%Wavelet tree construction of 
%
%Wavelet tree construction of multiscale wavelet coefficients, where the arrows indicate the parent-child relation between different levels of wavelet coefficients.
% \edward{Updating the font size and add text label ``root'' on the left hand side outside $C_0$.}
}
\label{fig:detail_coeff}
\end{figure}

\paragraph{Subband Coefficient Filtering.}
Based on the observations, we design the subband coefficient filtering procedure to locate and pack information-rich coefficients when building our representation.
% the following procedure to encode 3D shapes.
%
First, we keep all the coefficients in $C_0$, take them as the {\em coarse component\/} in our representation, and exclude all the coefficients in $D_2$, following observations (iii) and (iv).
Second, $C_0$ alone is insufficient to capture the details; compare Figure~\ref{fig:gt_recon} (d) vs.~(e).
We need the coefficients in $D_0$ and $D_1$.
However, simply including all coefficients in $D_0$ and $D_1$ will lead to a bulky representation.
Hence, we aim for a compact representation that can retain details by following observations (i) and (ii) to exploit the coefficient relations in $D_0$ and $D_1$.

Procedure-wise, since the subbands of $D_0$ share the same resolution and are positively correlated, as analyzed in observation (iii), we collectively examine the coefficient locations in all subbands of $D_0$ together.
For each coefficient location, we examine the sibling coefficients in $D_0^{LH}$, $D_0^{HL}$, and $D_0^{HH}$, selecting the one with the largest magnitude. We consider its magnitude value as the measure of \textit{information} for that coefficient location.
Next, we filter out the top K coefficient locations with the highest information content (refer to Figure~\ref{fig:detail_coeff} on the left) and store their location coordinates and associated coefficient values in $D_0$, along with their children's coefficient values in $D_1$. This forms the \textit{detail component} in our wavelet-tree representation, as illustrated in Figure~\ref{fig:detail_coeff} on the right.
Together with the coarse component, i.e., $C_0$, we construct our wavelet-tree representation.
Despite excluding all the coefficients in $D_2$ and selecting the top K coefficients in $D_0$, our representation can effectively achieve an impressive mean IOU of 99.56\%.

To efficiently process millions of 3D shapes, we utilize our wavelet-tree representation for encoding each shape, subsequently storing the results in the cloud. This constitutes a one-time preprocessing step. 
%
% Remarkably, our representation requires only a few megabytes per shape (4.6 MB), in stark contrast to the ground truth $256^3$ SDF, which consumes 67.1 MB.  
%
This aspect is particularly crucial for large-scale training, as it results in a 44.5\% reduction in both data streaming and loading, a significant improvement over directly using the $256^3$ SDF, as pointed out in the introduction. 
Moreover, we deliberately avoid using more sophisticated compression techniques due to the decompression overhead they incur, which can significantly slow down the model training.

\section{Diffusible Wavelet-tree Representation}

Next, we develop a representation that can be 
%easily
effectively trained and generated by a diffusion-based generative model.
This diffusion model is based on the DDPM framework \cite{ho2020denoising},  which formulates the generative process as a Markov chain. This chain comprises two key processes: (i) a forward process, which incrementally introduces noise into a data sample $x_0$ over $T$ time steps, eventually transforming it into a unit Gaussian distribution, denoted as $p(x_T) \sim N(0, I)$; and (ii) a reverse process, which is characterized by a generator network $\theta$ tasked to progressively remove noise from a noisy sample. In our approach, the generator network is designed to predict the original diffusion target $x_0$ directly from the noisy sample $x_t$, expressed as $f_\theta(x_t, t) \simeq x_0$. To achieve this, we employ a mean-squares loss objective.

%%%%%%%%%%%%%%%%%%%%%%%%%%%%%%%%%%%%%%%

\vspace*{-3mm}
\paragraph{Challenges.}
Our wavelet-tree representation, while being compact and efficient for data streaming, it encounters specific challenges during training. This representation consists of a coarse component, $C_0$, structured as a grid, and a detail component containing three irregular arrays. The detail component is derived from $D_0$ and $D_1$, as illustrated in Figure~\ref{fig:detail_coeff} (right).
A straightforward approach is to directly treat this representation as the diffusion target 
%$x_0$ 
and predict the coarse and detail components using a two-branch network.
However, 
%this approach requires 
it is hard to accurately predict the detail coefficient locations while balancing multiple objectives. 
We empirically observed that this approach struggles with convergence and leads to the collapse of model training.

\begin{figure}[t]
	\centering
	\includegraphics[width=1.0\columnwidth]{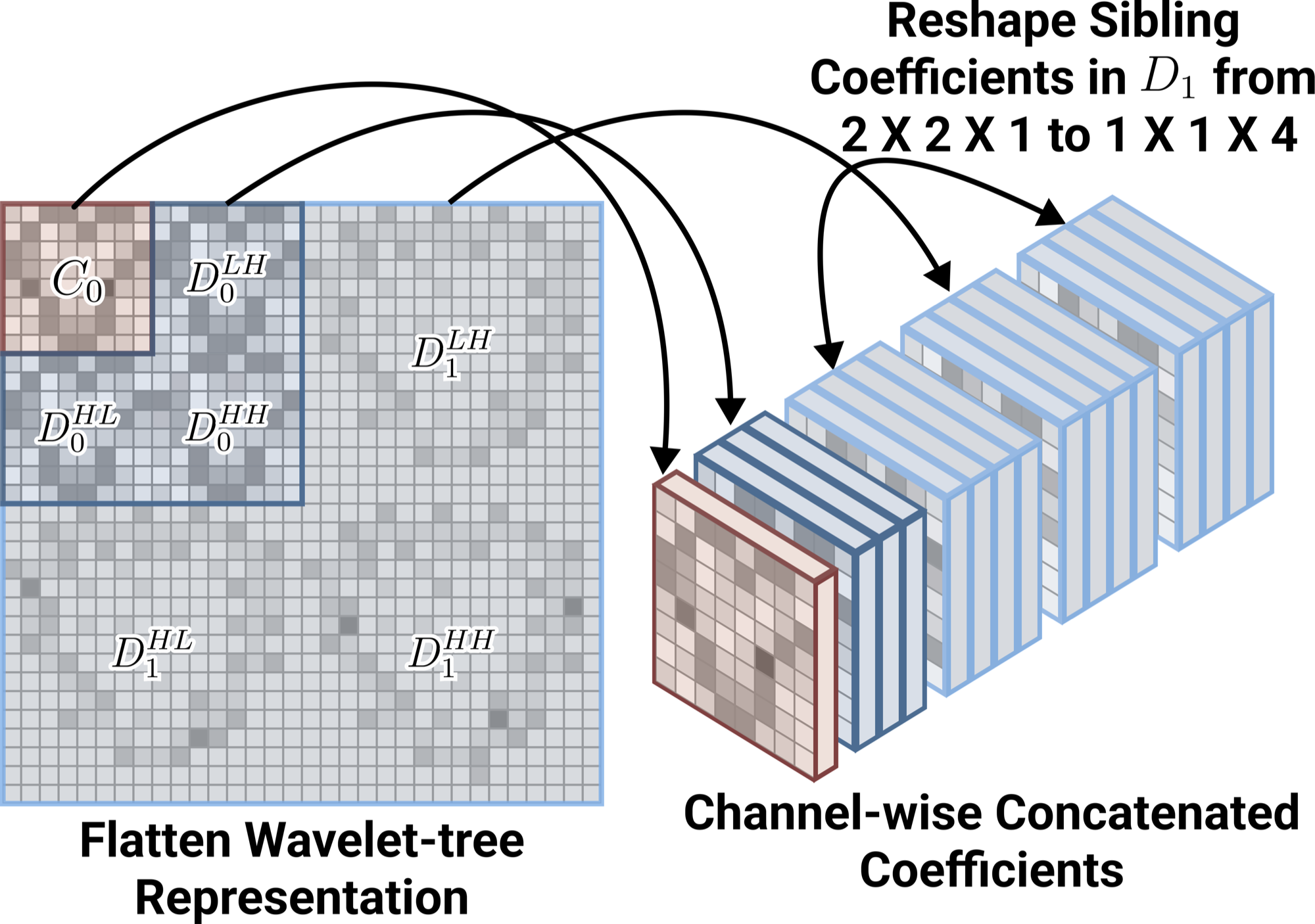}
	\caption{
 %To obtain a 
 Diffusible wavelet representation.
 First, we unpack and flatten the coefficients in our wavelet-tree representation (left).
 Following observation (iii), 
 %to further enhance the efficiency, 
 we 
 %identify correlated coefficients and 
 channel-wise concatenate sibling coefficients to reduce the spatial resolution (right).
 Here we concatenate each coefficient in $C_0$ with its three children in $D_0$ and the reshaped descendants in $D_1$ (each of size $1$$\times$$1$$\times$$4$).
 % \phil{{\color{red}{TODO}}: 
 % (i) Instead of saying ``spatially-concatenated coefficients'', please follow what we say in the main text, i.e., ``Flattened wavelet-tree representation'';
 % (ii) For ``Channel-wise concatenated coefficients'' in the figure, shall we consistently use lowercase c for concatenated and coefficients?
 % (iii) Also inside the figure image, need to add ``in $D_1$'' after the words ``sibling coefficients''}
 %
% \phil{1. add math symbols $C_0$, $D_0$ and $D_1$ appropriately and make the text labels large enough for visibility
%  2. don't just say reshape, ``reshape $2\times2$ to $1 \times 1 \times 4$ channels''?
%  3. please also mention so more clearly in figure caption.}
}
\label{fig:diffusible_wavelet}
\end{figure}

Another approach we tried is to flatten the extracted coefficients in our representation to avoid the irregularities in the detail component. 
As Figure~\ref{fig:diffusible_wavelet} (left) shows, we first arrange the coarse component $C_0$ at top left, then pack $D_0$ and $D_1$ successively around $C_0$ by arranging the extracted detail coefficients at their respective locations in each subband while leaving the remaining locations as zeros.
In this way, the input representation becomes a regular 2D grid for the DDPM to model.
However, this representation is spatially very large. The current U-Net architecture, widely adopted by existing diffusion models, creates GPU memory-intensive feature maps, which can lead to out-of-memory issues and result in low computational intensity, thus leading to poor utilization of the accelerator~\citep{hoogeboom2023simple}. Consequently, model training remains intractable.

%%%%%%%%%%%%%%%%%%%%%%%%%%%%%%%%%%%%%%%

\vspace*{-3mm}
% \paragraph{Diffusible Wavelet-tree Representation.}
\paragraph{Subband Coefficients Packing.}
To address these challenges, we draw inspiration from recent work on efficient 2D image generation~\cite{hoogeboom2023simple}. Our motivation is further bolstered by observation (iii), which highlights the relationship between coefficients. This insight enables us to effectively pack sibling subband coefficients, exhibiting similar structures (as illustrated in $D_0$ and $D_1$ of Figure~\ref{fig:wavelet_tree}), along with their corresponding children. These children are reshaped into a $1 \times 1 \times 4$ format, as demonstrated in Figure~\ref{fig:diffusible_wavelet} (right), and integrated into the channel dimension.
In this approach, the resulting representation (the diffusion model target) adopts a grid structure with reduced spatial resolution but an increased number of channels. Hence, this allows us to circumvent the use of memory-intensive feature maps, avoiding out-of-memory issues and enabling more efficient computation when training on millions of 3D shapes. Using this strategy in our 3D representation can approximately lead to a cubic-order speedup and a significant reduction in GPU memory usage, estimated to be around 64x, when applied to the same network architecture.

\section{Subband Adaptive Training Strategy}
\label{sec:adaptive_train}
In this step, our goal is to effectively train our diffusion model on this diffusible wavelet-tree representation. A key challenge for our model is ensuring that it can proficiently generate both the coarse and detailed components of the representation. An obvious choice for this purpose would be to employ a standard mean squared error (MSE) loss on the coefficient set $X$:
\begin{equation}
\label{eq:loss_simple} 
   L_{\text{MSE}}(X):= \frac{1}{|X|}\sum_{x_0}||f_\theta(x_t, t) - x_0||^2 \ , x_0 \in X, 
\end{equation}
where $x_t$ is the noised coefficient of $x_0$ at time step $t$.
Empirically, we observed that naively applying MSE to all coefficients results in significant quality degradation during training, as demonstrated in our ablation studies. We attribute this to two primary factors. Firstly, there is an imbalance in the number of coefficients across different scales. In the 2D case, the ratio of $|C_0|:|D_0|:|D_1|$ is $1:3:12$, and in the 3D case, it is $1:7:56$. Consequently, using a uniform MSE loss tends to disproportionately emphasize the fine-detail coefficients, even though the core shape information is more densely represented in $C_0$ than in $D_0$ and $D_1$.
Secondly, the majority of the detail coefficients are close to zeros, with only a few having high magnitude or \textit{information} which contain maximum high-frequency information about the shape. Therefore, uniformly sampling a loss across these coefficients can result in a sub-optimal training mechanism due to the imbalance in the number of high-magnitude detail coefficients.

An initial approach to tackle the issue involves defining three separate MSE losses for $C_0$, $D_0$, and $D_1$, then combining these losses with equal weights. However, this approach still treats the losses on detail coefficients uniformly, without resolving the imbalance issue due to sparsity in the detail subbands.
This oversight is empirically evidenced by the subpar performance observed in our ablation study.

\vspace*{-3mm}
\paragraph{Subband Adaptive Training.}
To address these issues, we develop a subband adaptive training strategy. This approach is specifically designed to focus more effectively on the high magnitude detail coefficients while still maintaining a balanced consideration for the other remaining detail coefficients. This ensures that they are not completely overlooked.
Specifically, for each subband in $D_0$, say $D_0^{LH}$, we first locate the coefficient of the largest magnitude in $D_0^{LH}$.
Denoting $v$ as its magnitude value, we then identify all coefficients in $D_0^{LH}$ with magnitude larger than $v/32$; we regard these coefficients as important, and record their spatial locations into coordinate set $P_0^{LH}$.
Similarly, we can obtain coordinate sets $P_0^{HL}$ for $D_0^{HL}$ and $P_0^{HH}$ for $D_0^{HH}$.
As sibling coefficients are positively-correlated, we union the three coordinate sets into coordinate set $P_0$, which records the spatial locations of the important detail coefficients.

On the other hand, we define $P'_0$ as the coordinate set complement to $P_0$ in the spatial domain of the $D_0$ subband.
Using $P_0$ and $P'_0$, we can then formulate the training loss as
\begin{equation}
L_{\text{MSE}}(C_0)
+
\frac{1}{2}
\Big[
%L_{\text{MSE}}(P_0)
\sum_{D}L_{\text{MSE}}(D[P_0])
+
%L_{\text{MSE}}(P_0')
\sum_{D}L_{\text{MSE}}(R(D[P_0']))
\Big],
\end{equation}
where $D$ is a subband in $\{D_0, D_1\}$;
$D[P]$ denotes $D$'s coefficients at locations in $P$;
and
$R$ is a function that randomly picks some of the coefficients in $D[P_0']$.
Importantly, the number of coefficients picked by $R$ is $|P_0|$, such that we can balance the last two terms in our loss with the same number of coefficients.
Note that using random sampling can effectively regularize the less important small coefficients while not completely ignoring them in the training.
% \edward{
Also, $P_0$ includes far less coefficients than the whole domain of $D_0$ (only $\sim 7.4\%$), so the model training can effectively focus on the crucial information while attending to the details.

\vspace*{-3mm}
\paragraph{Efficient Loss Computation.}
To calculate the losses, we may store $P_0$ and $P'_0$ as irregularly-sized coordinate sets and employ slicing to process the generation target and network prediction. Doing so is, however, highly inefficient, as it requires distinct computations for different data samples and prevents us from leveraging code compilation features for training speedup.

To improve the model training efficiency, we propose to use a fixed-size binary mask to represent the coordinate set, where a value of one indicates the associated location as selected. 
The MSE loss can then be efficiently calculated by masking both the generation target and network prediction. 
This approach eliminates the need for irregular operations and allows for efficient use of code compilation in PyTorch to accelerate the training.

\section{Extension for Conditional Generation}
\label{sec:conditional_generation}
\begin{figure}[t]
	\centering
	\includegraphics[width=1.0\columnwidth]{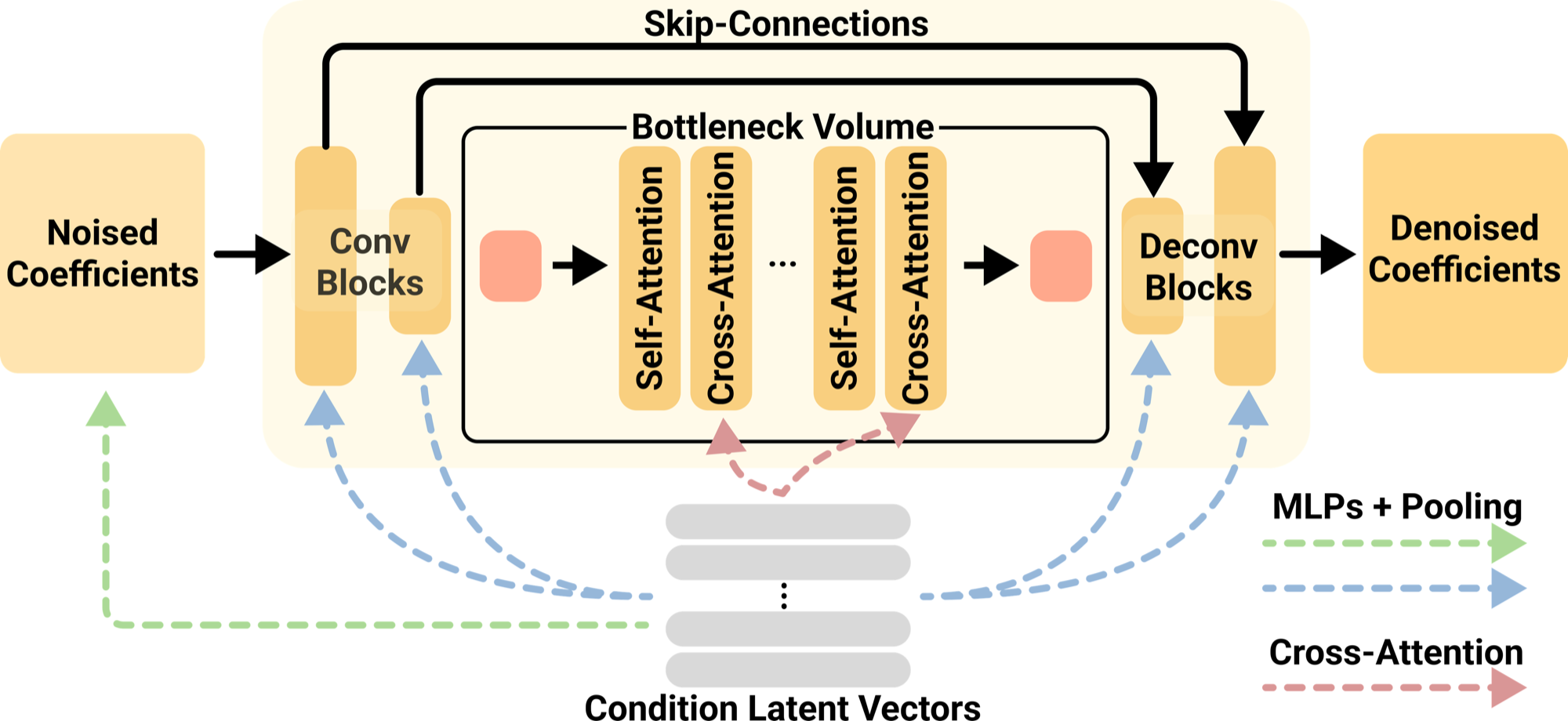}
	\caption{
 Our generator network progressively downsamples input coefficients to a bottleneck feature volume (middle). 
This volume goes through attention layers and deconvolution for upsampling to predict the denoised coefficients.
If the condition latent vectors are available, we simultaneously transform these vectors and adopt them at three locations in our architecture:
(i) concatenating with the input noised coefficients (the green arrow);
(ii) conditioning the convolution and deconvolution blocks (the blue arrows); and
(iii) cross-attention with the bottleneck volume (the red arrows).
}
\label{fig:network_architecture}
\end{figure}
.
Our framework is versatile and can be extended beyond unconditional generation to accommodate conditional generation across various modalities. To achieve this, we adopt different encoder for each modality that  transforms a given condition into a sequence of latent vectors. Subsequently, these vectors are injected into the generator using multiple conditioning mechanisms. We also use a classifier-free guidance mechanism~\cite{ho2022classifier}, which has empirically demonstrated greater effectiveness in conditional settings.

\para{Condition Latent Vectors.}
We deliberately convert all input conditions into a sequence of latent vectors, which we call \textit{condition latent vectors}, to preserve the generality of our conditioning framework. This approach eliminates the need to devise new specific condition mechanisms to diffusion model for each modality, thereby enabling our framework to function seamlessly across various modalities. Our encoder for different modality are described below:
\begin{enumerate}[label=(\roman*)]
\item
%(i) 
{\em Single-view image.}
Given a rendered image of a 3D model, we utilize the pre-trained CLIP L-14 image encoder \citep{radford2021learning}  to process the image. The latent vectors extracted from just before the pooling layer of this encoder are then used as the conditional latent vectors.
\item
%(ii)
{\em Multi-view images.}
%In this scenario, we 
We are provided with four images of a 3D model, each rendered from one of 55 predefined camera poses (selected randomly). To generate the conditional latent vectors, we first use the CLIP L-14 image encoder to process each rendered image individually to produce an image latent vector. Considering the camera poses, we maintain 55 trainable camera latent vectors, each corresponding to one camera pose and matching the dimensionality of the latent vectors encoded by the CLIP image encoder. For each encoded image latent vector, we retrieve the corresponding trainable camera latent vector based on the camera pose of the image. This camera vector is then added to each image latent vector in the sequence in an element-wise fashion. Finally, the four processed sequences of latent vectors are concatenated to form the conditional latent vectors.
\item
%(iii)
{\em 3D point cloud.} 
We utilize three Multi-Layer Perceptron (MLP) layers to first transform the given point cloud into feature vectors like PointNet \cite{qi2017pointnet}. These vectors are then aggregated using the PMA block form the Set Transformer layer~\cite{lee2019set}, resulting in sequnece of latent vectors that serve as the condition.
% \edward{Set transformer is a method}
% \phil{any s?} \edward{Yes}
%
\item
%(iv)
{\em Voxels.} 
We utilize two 3D convolution layers to progressively downsample the input 3D voxels into a 3D feature volume. This volume is subsequently flattened to form the desired conditional latent vectors.
\end{enumerate}

\para{Network Architecture.}
Figure~\ref{fig:network_architecture} illustrates the network architecture of our generator. The main branch, highlighted by yellow boxes, adopts a U-ViT architecture~\cite{hoogeboom2023simple}.  The network uses multiple ResNet convolution layers  for downsampling our noised coefficients into a feature bottleneck volume, as shown in the middle part of Figure~\ref{fig:network_architecture}. Following this step, we apply a series of attention layers to the volume. The volume is then upscaled using various deconvolution layers to produce the denoised coefficients. A key feature of our architecture is the inclusion of learnable skip-connections between the convolution and deconvolution blocks, which have been found to enhance stability and facilitate more effective information sharing~\cite{huang2023scalelong}. 

Moreover, when condition latent vectors are available, we integrate them into our generation network at three distinct locations in the U-ViT architecture, as depicted in the bottom part of Figure~\ref{fig:network_architecture}. Initially, these latent vectors are processed through MLP layers and a pooling layer to yield a single latent vector (highlighted by the green arrow in the left section of Figure~\ref{fig:network_architecture}). This vector is subsequently concatenated as additional channels of the input noise coefficients.
Second, following a similar process, we convert condition latent vectors to another latent vector. However, this vector is utilized to condition the convolution and deconvolution layers via modulating the affine parameters of group normalization layers~\cite{dhariwal2021diffusion}. This integration is represented by the blue arrows in Figure~\ref{fig:network_architecture}.
Lastly, to condition the bottleneck volume, an additional positional encoding is applied to the condition latent vectors in an element-wise fashion. These vectors are then used in a cross-attention operation with the bottleneck volume, as indicated by the red arrows in Figure~\ref{fig:network_architecture}.

\section{Results}
%\section{Experiments}
\label{sec:results}

In this section, we begin by presenting the experimental setup, followed by both quantitative and qualitative results obtained using various input conditions. Further, we showcase that our generative model is adaptable to shape completion tasks without additional training. Last, we present comprehensive analysis and ablations on our framework.
% Additional qualitative results are available in the supplementary materials.
\subsection{Experimental Setup}
\label{sec:exp_setup}
\para{Dataset.}
%In our experiments, 
We compile a new very large-scale dataset, consisting of more than 10 million 3D shapes, from 18 existing publicly-available sub-datasets:
% \edward{
%
% Philip: note that I just sort them in chron. order
ModelNet~\cite{vishwanath2009modelnet},
ShapeNet~\cite{chang2015shapenet},
SMLP~\cite{loper2015smpl},
Thingi10K~\cite{zhou2016thingi10k},
SMAL~\cite{zuffi2017smal},
COMA~\cite{ranjan2018coma},
House3D~\cite{wu2018building},
ABC~\cite{koch2019abc},
Fusion 360~\cite{willis2021fusion},
3D-FUTURE~\cite{fu20213d},
BuildingNet~\cite{selvaraju2021buildingnet},
DeformingThings4D~\cite{li20214dcomplete},
FG3D~\cite{liu2021fine},
Toys4K~\cite{stojanov2021using}, 
ABO~\cite{collins2022abo},
Infinigen~\cite{raistrick2023infinite},
Objaverse~\cite{deitke2023objaverse},
and two subsets of ObjaverseXL~\cite{deitke2023objaverse} (Thingiverse and GitHub).
Among the sub-datasets, some contain specific object classes, \textit{e.g.}, CAD models (ABC and Fusion 360), furniture (ShapeNet, 3D-FUTURE, ModelNet, FG3D, and ABO), humans (SMLP and DeformingThings4D), animals (SMAL and Infinigen), plants (Infinigen), faces (COMA), houses (BuildingNet and House3D), etc. Beyond these, the Objaverse and ObjaverseXL datasets encompass generic objects collected from the internet, thereby not only covering the aforementioned categories but also offering a more diverse range of objects.
For the data split, we randomly divided each sub-dataset into two parts: a training set consisting of \(98\%\) of the shapes and a testing set comprising the remaining \(2\%\). The final train and test sets were then compiled by combining the respective train and test sets from each sub-dataset.

For each shape in our dataset, we generate a TSDF and its wavelet-tree representation for model training.
On the other hand, we prepare various additional inputs for the conditional generation tasks.
For image inputs, we randomly sampled 55 pre-defined camera poses and rendered 55 images for each object according to these poses, using the scripts provided by~\cite{jun2023shap}. For voxel inputs, we prepared two sets of voxels at different resolutions (\(16^3\) and \(32^3\)) for each 3D object and trained separate models for each resolution. Lastly, we randomly sampled 25,000 points on the surface of each 3D shape to generate the point cloud input.

\para{Training Details.}
We train our shape model using the Adam Optimizer~\cite{kingma2014adam}  with a learning rate of 1e-4 and a batch size of 96. 
%
%We use a batch size of 96 for training our models.
To stabilize the training, we employ an exponential moving average with a decay rate of 0.9999 in the model updates, in line with existing 2D large-scale diffusion models~\cite{rombach2022high}. 
Our model is trained on 48 $\times$ A10G with 2M-4M iterations, depending on the input condition. 
\begin{figure*}[t]
	\centering
	\includegraphics[width=0.9\linewidth]{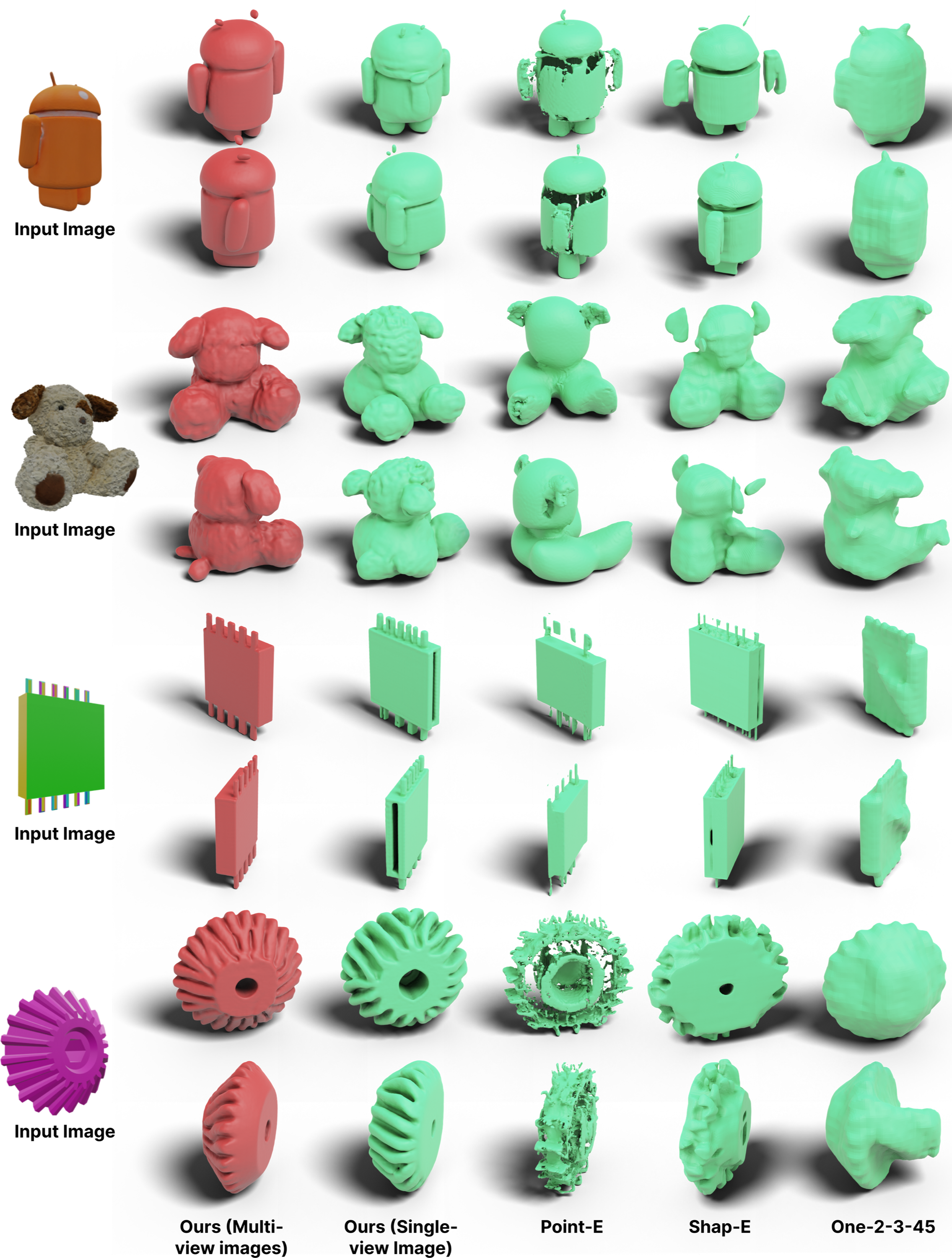}
	\caption{
Visual comparisons for the Image-to-3D generation task reveal that our method outperforms three major generative models: Point-E~\cite{nichol2022point}, Shap-E~\cite{jun2023shap}, and One-2-3-45~\cite{liu2023one}. Our single-view model generates more accurate shapes compared to these baselines, and the multi-view model further enhances shape fidelity with additional view information.
 }
\label{fig:image_3d_comp}
\end{figure*}

\para{Evaluation Dataset.}
For qualitative evaluation, we provide visual results based on the inputs in the unseen test set of our compiled large-scale dataset. 
For quantitative evaluation, we prepare two evaluation sets for metric computation.
In particular, we randomly select 50 shapes from the test set of each sub-dataset to form the first evaluation set for computing the metrics. We refer to this dataset as ``Our Val'' dataset throughout the remainder of the paper.
We also utilize the Google Scanned Objects (GSO) dataset to create an additional evaluation set to evaluate the cross-domain generalization capability for our method, noting that our model has not been trained on this dataset.
Please note that while a subset of the Google Scanned Objects (GSO) dataset has been used as an evaluation set in One-2345~\cite{liu2023one}, we have included all objects from this dataset in our study to ensure a more comprehensive evaluation.

\begin{figure}[t]
	\centering
	\includegraphics[width=1.0\linewidth]{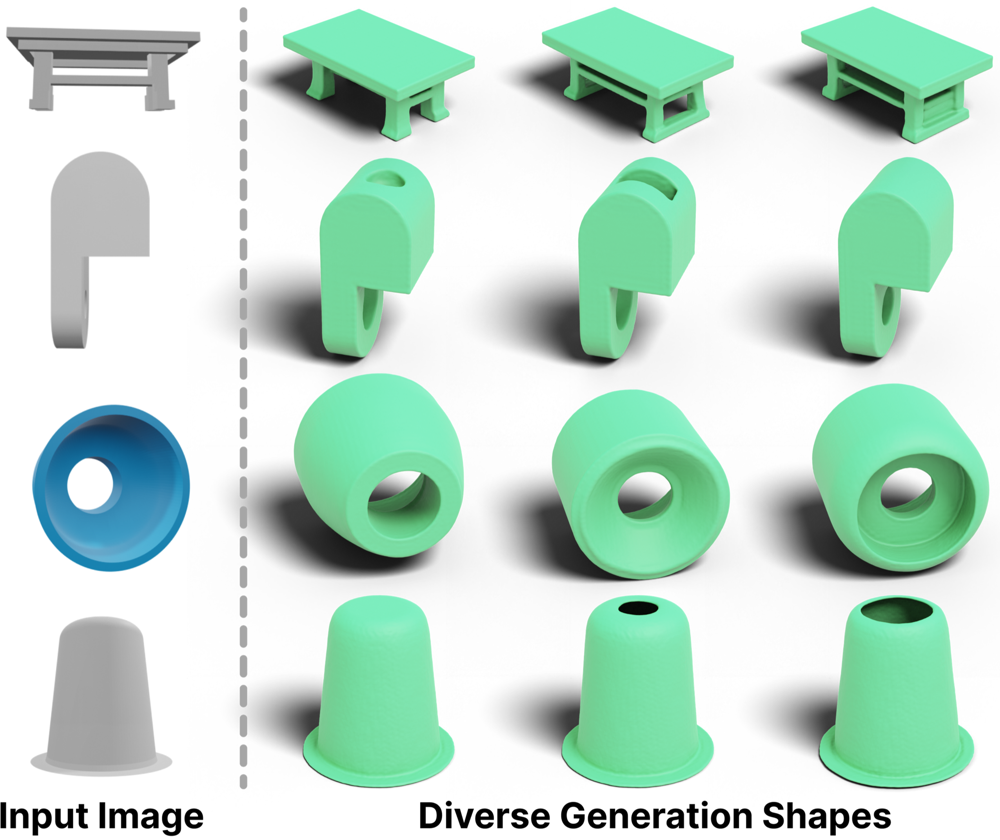}
	\caption{ Our model demonstrates the capability to generate varied results from a single input image, accurately resembling the visible portions while offering diversity in unseen areas.}
\label{fig:diversity_result}
\end{figure}

\begin{figure*}[t]
	\centering
	\includegraphics[width=1.0\linewidth]{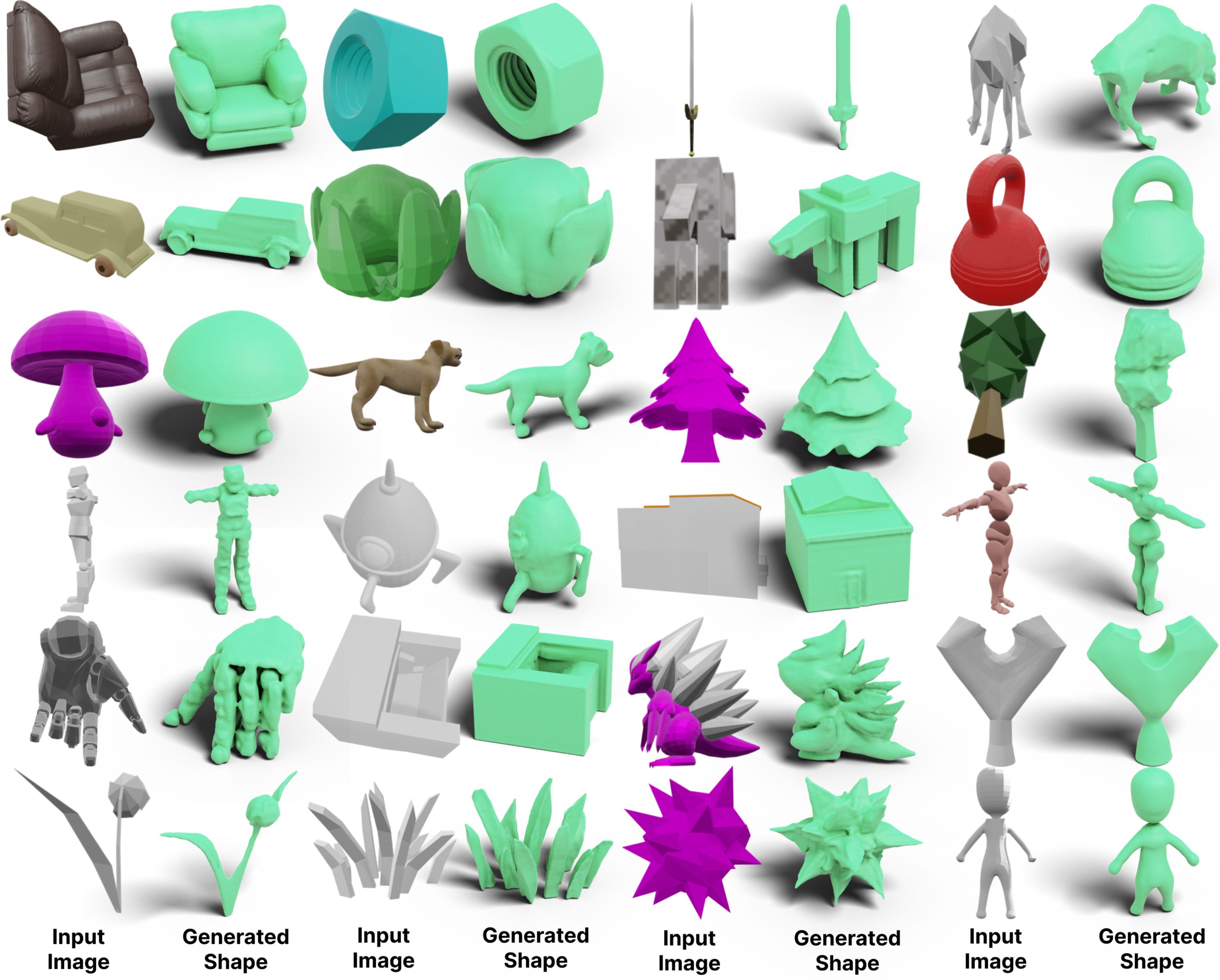}
	\caption{
Our single-view conditioned generation model yields a wide variety of shapes. Our model adeptly generates both CAD objects like screws, chairs, and cars, as well as organic forms such as humans, animals, and plants.
 %  \phil{1. Use less saturated colors. Just photoshop the image.
 % 2. Do you think we need text labels or explanation to aid readers which is which in the figure?}
 }
\label{fig:single_view_gallery}
\end{figure*}

\para{Evaluation Metrics.}
In the conditional task, we evaluate performance by comparing the similarity between the generated shape and the associated ground-truth shape using two metrics:
(i) Intersection over Union (IoU), which measures the volume ratio of the intersection to the union between the voxelized volumes; and
(ii) Light Field Distance (LFD)~\cite{chen2003visual}, which assesses the similarity between two sets of images rendered from various viewpoints.
%
% For the details on the metrics, please refer to the supplementary materials.

For the unconditional task, we adopt the method described in~\cite{zhang20233dshape2vecset} to evaluate the generation performance using the Frechet Inception Distance (FID)~\cite{heusel2017gans}.
% For the unconditional task, we follow~\cite{zhang20233dshape2vecset} to evaluate the Frechet Inception Distance (FID) for the generation performance.
%
In particular, we render an image for each shape from Our Val set. Following this, we apply the same rendering process to a generated set of shapes of equivalent size.  We then compute a feature vector for each rendered image and assess the difference in the distribution of these feature vectors between the two sets as the final metric.

%%%%%%%%%%%%%%%%%%%%%%%%%%%%%%%%%%%%%%%%%%%%%%%%
% \subsection{Image-to-3D Generation}
\begin{table}
\centering
{\small
\caption{
Quantitative evaluation of the Image-to-3D task shows that our single-view model excels the baslines, achieving the highest IoU and lowest LFD metrics. Incorporating additional information, our multi-view model further enhances performance.} % 
\label{tab:imag_to_3d_recon}
% \begin{tabular}{c|c|cccc}
% \toprule
% \multirow{2}{*}{Setting} & \multirow{2}{*}{Methods} & \multicolumn{2}{c}{GSO Dataset} & \multicolumn{2}{c}{TODO XL} \\
%                    &                    & \textbf{LFD $\downarrow$}         & \textbf{IOU $\uparrow$}         & \textbf{LFD $\downarrow$}         & \textbf{IOU $\uparrow$}         \\     
% \midrule
% \multirow{4}{*}{Single-view}  & Point-E \citep{nichol2022point} & 5018.73&0.1948 &6181.97 &0.2154 \\
%                    & Shap-E \citep{jun2023shap} & 3824.48 & 0.3488 & 4858.92 & 0.2656\\
%                    & One-2-3-45 \citep{liu2023one} & 4414.03 & 0.3020 & 5123.98 & 0.2036\\
%                    & Ours  & 3406.61 &	0.5004 &	4071.33 &	0.4285\\
% \midrule
% Multi-view &Ours & 1890.85	& 0.7460 &	2217.25 & 0.6707 \\
% \bottomrule
\begin{tabular}{c|cccc}
\toprule
\multirow{2}{*}{\textit{Method}}   & \multicolumn{2}{c}{GSO Dataset}  & \multicolumn{2}{c}{Our Val Dataset}  \\
& LFD $\downarrow$ & IoU $\uparrow$ & LFD $\downarrow$ & IoU $\uparrow$ \\
\midrule
Point-E \citep{nichol2022point} & 5018.73 & 0.1948 & 6181.97 & 0.2154\\
Shap-E \citep{jun2023shap} & 3824.48 & 0.3488 & 4858.92 & 0.2656 \\
One-2-3-45 \citep{liu2023one} & 4397.18 & 0.4159 & 5094.11 & 0.2900  \\
Ours (Single view)  & \textbf{3406.61} &	\textbf{0.5004} &	\textbf{4071.33} &	\textbf{0.4285}\\ 
\midrule
Ours (Multi view) & 1890.85	& 0.7460 &	2217.25 & 0.6707\\
\bottomrule
\end{tabular}
}
% \caption
\end{table}
 \subsection{Quantitative Comparison with Other Large Generative Models}

In this experiment, we contrast our method with other large image-to-3D generative models.  Our analysis encompasses two distinct model settings: single-view and multi-view. The single-view model uses a solitary image, which serves as an input for our wavelet generative model. 
In cases where multiple images are accessible, our multi-view model comes into play. This model uses four images along with the camera parameter as the condition.

We present the quantitative results in Table~\ref{tab:imag_to_3d_recon} and the qualitative comparison results in Figure~\ref{fig:image_3d_comp}. As shown in Table~\ref{tab:imag_to_3d_recon}, our single-view model significantly outperforms all baseline models by a considerable margin in both the IoU and LFD metrics. It is noteworthy that LFD, being a rotation-insensitive metric, indicates that our results do no depend on the alignment between the generated shapes and the ground-truth shapes. Additionally, Figure \ref{fig:image_3d_comp} reveals that our method captures not only global structures but also fine local details, as well as more intricate geometric patterns, compared to the baselines. This is particularly evident in rows 7-8 of Figure~\ref{fig:image_3d_comp}, showing that our method is able to more accurately capture the geometric patterns of lines than the other baselines. Furthermore, rows 3-4 showcase how our model effectively captures the rough coat pattern of the shape, further demonstrating its superiority. Finally, we would like to highlight that the geometry produced by our method features clean and smooth surfaces, in contrast to those generated by other baseline methods. 

On the other hand, when provided with additional three views, our model exhibits a substantial improvement in results. Both LFD and IoU metrics, as indicated in Table~\ref{tab:imag_to_3d_recon}, show that the generative model captures a greater extent of the global shape. This improvement is expected, since the network can access more information in the form of multiple views; however, four images still constitute a limited set of views for completely reconstructing a shape. Moreover, multiple views aid in better capturing local geometric details, as demonstrated in Figure~\ref{fig:image_3d_comp} rows 3-4 and 7-8. Also, it is worthy to mention recent advancements, such as Zero123~\citep{liu2023zero} and Zero123-XL~\citep{deitke2023objaverse}, which have the capability to generate multiple views from a single view. This advancement potentially allows our multi-view model to operate effectively with only a single view available since a single-view can be converted to a multi-view using Zero123 or Zero123-XL, after which our model can be applied. Yet, exploring this possibility remains a topic for future research.

We attribute this success to two major factors. Firstly, we posit that our wavelet-tree representation is almost lossless, as discussed in Table~\ref{tab:rep_compare}, compared to Point-E and Shape-E. This characteristic likely makes the upper bound of our representation much easier to capture using a diffusion model. Moreover, our adaptive training scheme enables our network to capture more local details and geometric patterns, by efficiently learning to model the information-rich coefficients in the detail coefficient subbands.

\subsection{Image-to-3D Generation}

\begin{figure*}[t]
	\centering
	\includegraphics[width=1.0\linewidth]{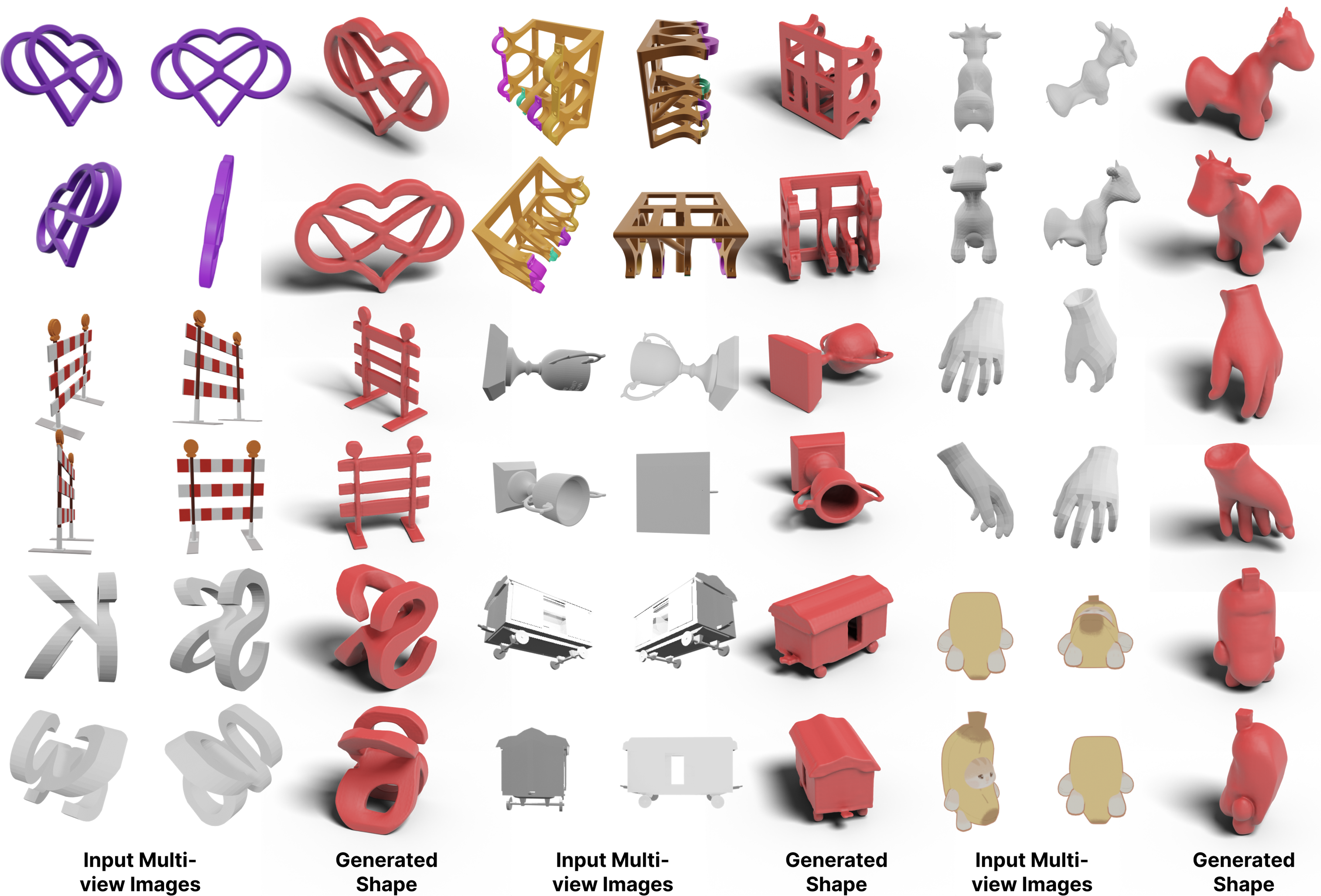}
	\caption{
Our multi-view conditioned generation model can utilize multi-view information to create diverse and coherent shapes with complex topologies, exemplified by the CAD objects in the first two rows.
 }
\label{fig:multi_view_gallery}
\end{figure*}

In this section, we present additional qualitative results on single-view conditioning from our val set. As depicted in Figure \ref{fig:single_view_gallery}, it is evident that our method can generate objects from a variety of categories. This includes CAD objects, such as the screw shown in row 1, furniture like the chair in row 5, and buildings exemplified by the house in row 4. Additionally, our method effectively represents organic shapes, including humans (as seen in rows 4 and 6), animals (with the dog in row 3), and plants (illustrated by the Christmas tree and mushroom in row 3).

Our model, being a generative one, can produce multiple variations for a given condition, as demonstrated in Figure \ref{fig:diversity_result} using single-view images. %Two key observations emerge from these results. First, the model can faithfully reconstruct the parts of the image that are visible.
In addition to the visible regions that can be faithfully reconstructed by our approach,
%Second, 
it further imaginatively reconstructs the parts that are invisible. For instance, in Figure \ref{fig:diversity_result}, rows 2 and 4, the top of the CAD object is invisible, leading the model to imagine it. 
This trade-off between adhering to visible parts and creatively interpreting invisible ones could be further explored by adjusting the classifier-free guidance weight. We leave this aspect for future work.

We also present additional results for the multi-view approach in Figure \ref{fig:multi_view_gallery}. Again, it demonstrates that our method is able to generate objects across diverse categories. Moreover, there is a noticeable alignment of the objects with the input images, which is more pronounced compared to the single-view approach. %This intuitively makes sense as we have access to more information. 
Our method is also capable of generating objects with highly complicated topologies and intricate geometries, as demonstrated by the CAD examples (see the second object in rows 1-2 and the first object in rows 5-6).

\subsection{Point-to-3D Generation}
\begin{figure*}[t]
	\centering
	\includegraphics[width=1.0\linewidth]{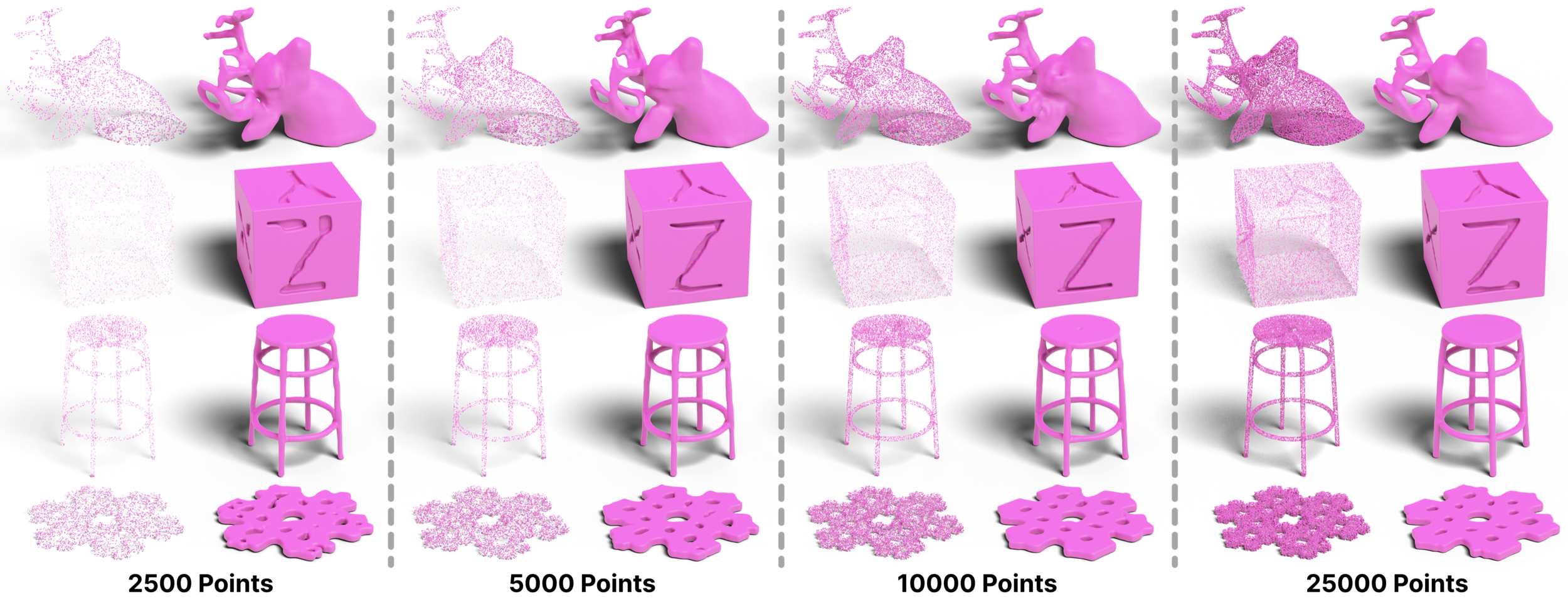}
	\caption{Visual comparisons, based on the number of input points, highlight our model's ability to robustly generate thin structures, like the deer horn or the chair leg, with a reasonable number of points ($\geq 5000$).}
\label{fig:point_comp}
\end{figure*}

\begin{table}
\centering
{\small
\caption{
The quantitative evaluation reveals that our model's performance is not significantly impacted by the number of points of inputs. Even with inputs of 5000 points, it manages to deliver reasonable reconstructions, though trained on 25000-point inputs.
} % 
\label{tab:point_comp}
\begin{tabular}{c|cccc}
\toprule
\multirow{2}{*}{Metrics} & \multicolumn{4}{c}{Number of Points}  \\
                         & 2500    & 5000    & 10000   & 25000   \\
\midrule
LFD $\downarrow$                      & 1857.84 & 1472.02 & 1397.39 & 1368.90 \\
IoU $\uparrow$                      & 0.7595  & 0.8338  & 0.8493  & 0.8535  \\
\bottomrule
\end{tabular}
}
% \caption
\end{table}

In this experiment, our objective is to take a point cloud as input and produce a TSDF following its geometry. Our encoder, comprising a PointNet \cite{qi2017pointnet} and a PMA block from the Set Transformer \cite{lee2019set}, is adept at handling a variable number of points. This versatility allows us to train the model with 25,000 points, while also giving us the flexibility to input an arbitrary number of points during testing.

We conduct an ablation study to assess how the quality of generation is influenced by different sets of points, as detailed in Table \ref{tab:point_comp}. Our findings reveal that an increase in the number of points leads to improved IoU results on our val set. %This result is intuitive since more points equate to finer detail. 
Notably, even with sparse point clouds with as few as 5,000 points, our model achieves a reasonable IoU.

\begin{figure*}[t]
	\centering
	\includegraphics[width=1.0\linewidth]{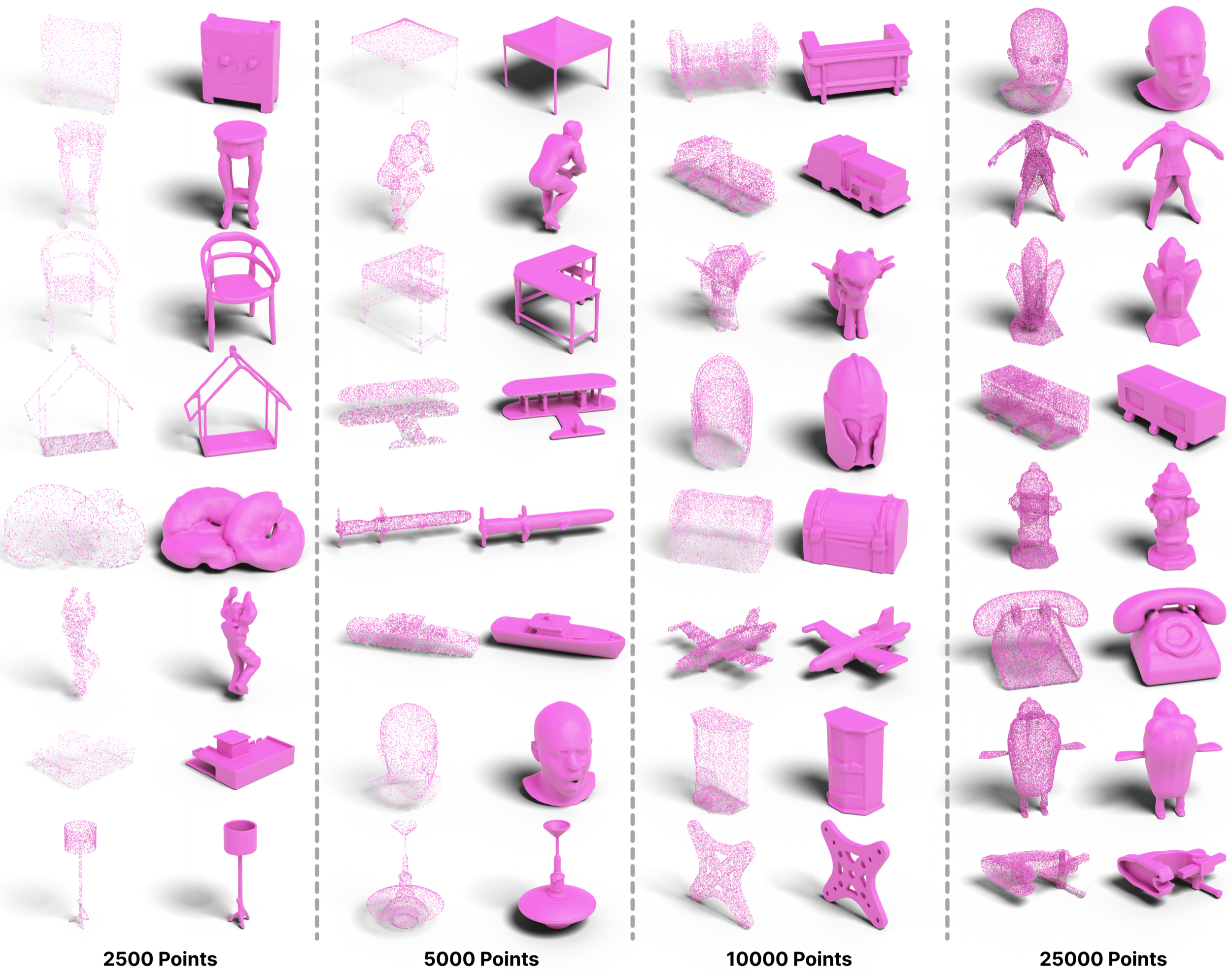}
	\caption{Our point cloud conditioned generation results demonstrate that our model is capable of producing shapes with intricate geometric details while preserving the complex topologies of the input point clouds.}
\label{fig:point_gallery}
\end{figure*}

This analysis is also visually represented illustrated in Figure \ref{fig:point_comp}. Here, we observe that certain details are lost when a lower number of points are used, as evident in row 2. However, it's worth mentioning that, in general, our method performs well even with fewer points.
We also present additional visual results in Figure~\ref{fig:point_gallery}. These results demonstrate that our method performs consistently well across various types of object and showcasing robustness predominantly to the number of points. 

\subsection{Voxel-to-3D Generation}
\begin{figure*}[t]
	\centering
	\includegraphics[width=1.0\linewidth]{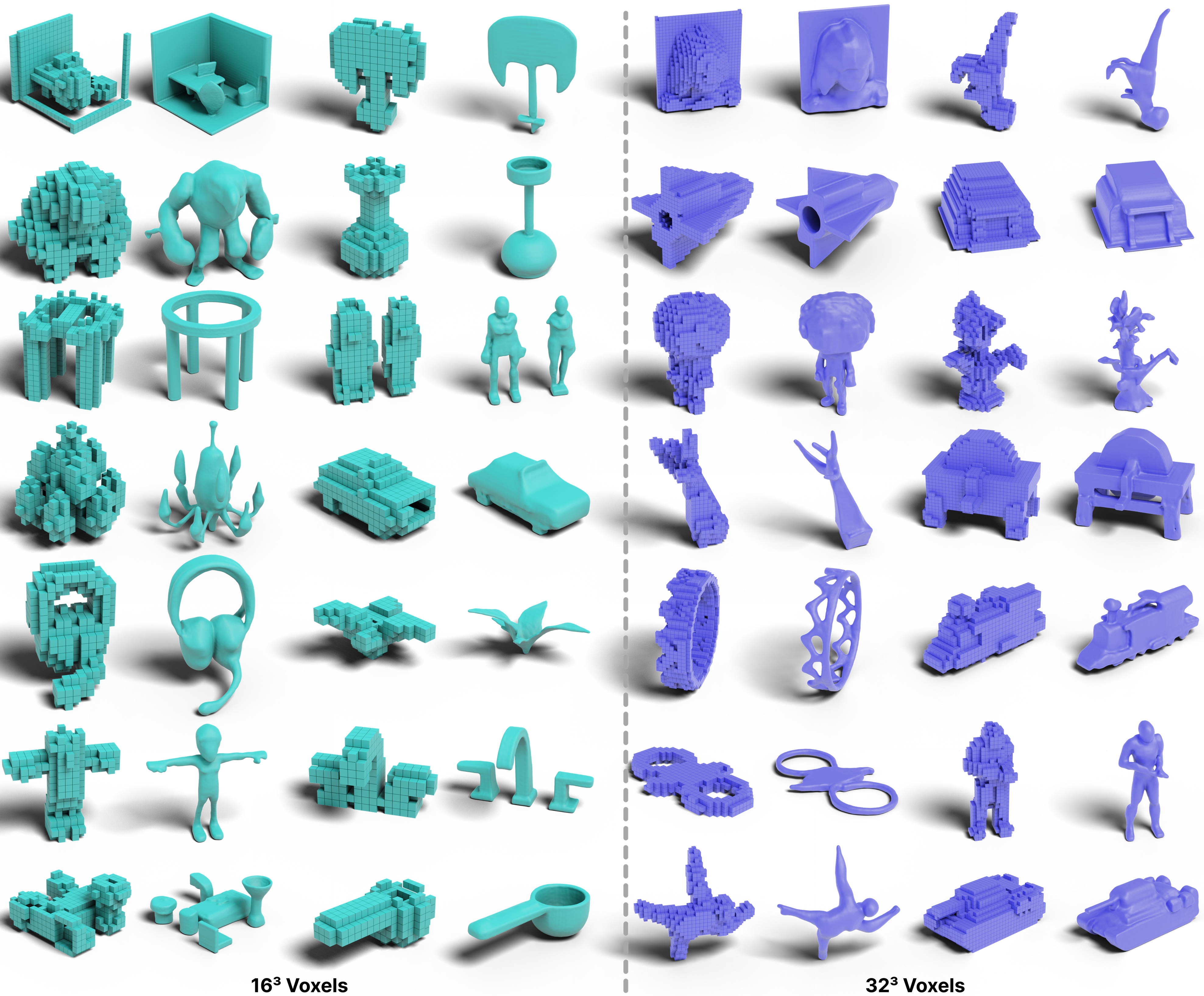}
	\caption{
Our voxel-conditioned generative model excels in creating high-quality outputs from low-resolution inputs, imaginatively introducing various plausible geometric patterns. This is exemplified by the creation of holes in the crown, which are not available in the initial inputs.
 }
\label{fig:voxel_gallery}
\end{figure*}

\begin{table}
\centering
{\small
\caption{
Our method is quantitatively compared with traditional voxel upsampling techniques, specifically nearest neighbour upsampling and trilinear interpolation, followed by marching cubes~\cite{lorensen1998marching} for mesh extraction. Our generative model significantly surpasses these two baselines in both Light Field Distance (LFD) and Intersection Over Union (IoU) metrics.} % 
\label{tab:voxel_comp_tab}
\begin{tabular}{c|c|cc}
\toprule
 
Setting                                                  & Methods                           & LFD $\downarrow$                            & IoU $\uparrow$                                                  \\
 \midrule
                                 & Ours      & 2266.41                        &  0.687 \\
 
                                 & Nearest                           & 6408.82                        & 0.2331                                               \\
\multirow{-3}{*}{Voxel ($16^3$)} & Trilinear                         & 6132.99                        & 0.2373                                               \\
 \midrule
                                 & Ours                              &  1580.98 &  0.7942                        \\
                                 & Nearest   & 3970.49                        & 0.4677                                               \\
\multirow{-3}{*}{Voxel ($32^3$)} & Trilinear & 3682.83                        & 0.4719 \\
\bottomrule
\end{tabular}
}
% \caption
\end{table}
\begin{figure}[t]
	\centering
	\includegraphics[width=1.0\linewidth]{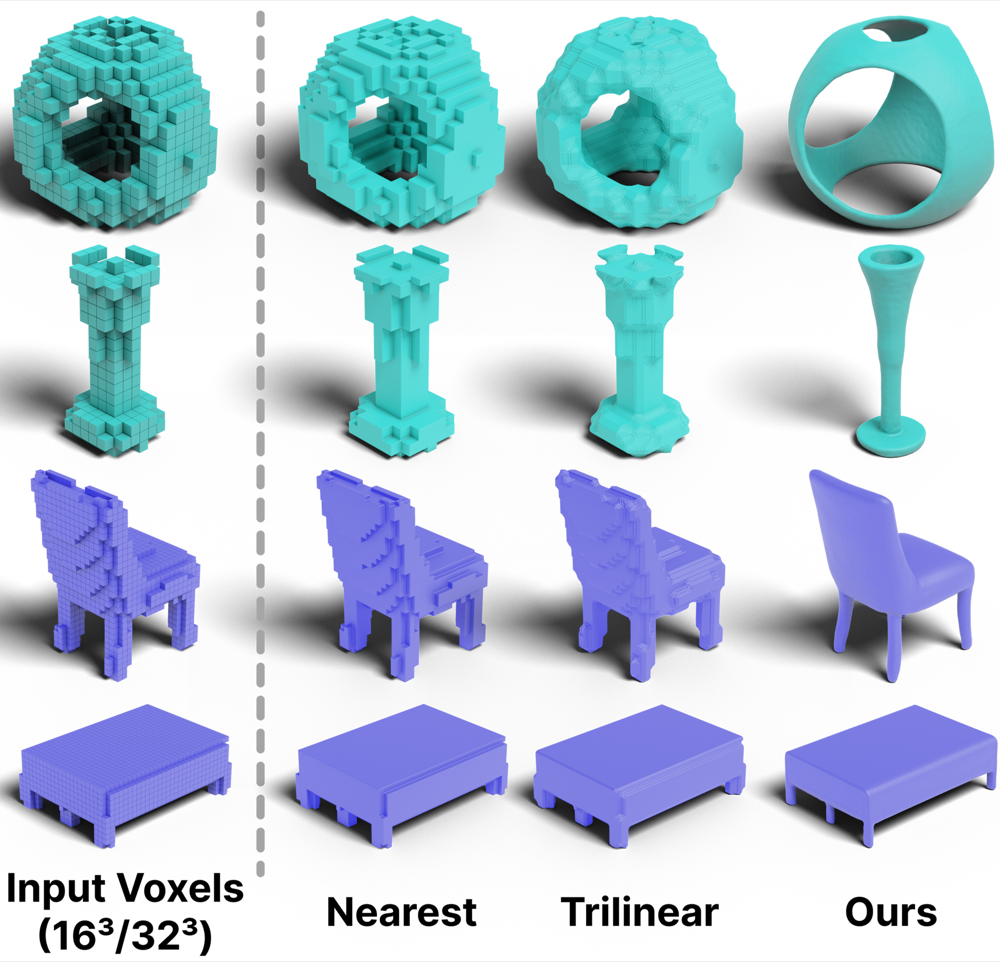}
	\caption{In comparison with meshes generated from interpolation using nearest neighbor upsampling and trilinear interpolation, our generation results display notably smoother surfaces.}
\label{fig:voxel_comp}
\end{figure}

% \subsection{Text-to-3D Generation}
Next, we explore the use of low-resolution voxels as input for our model, which outputs a Signed Distance Function following its geometry. We trained two distinct models on different voxel resolutions: \(16^3\) and \(32^3\). Both models employ the same encoder, which utilizes two 3D convolutional blocks to downsample the input into a conditional latent vector.

The qualitative results for both resolutions are displayed in Figure~\ref{fig:voxel_gallery}. From this analysis, we observe three main points.
First, our method successfully creates smooth and clean surfaces. Second, despite the ambiguity present in some examples for both voxel \(16^3\) (as seen in the human examples in row 3) and voxel \(32^3\) (as seen in the crown examples in row 5), our method produces plausible shapes. Finally, our method also performs well with disjoint objects (as demonstrated in the human examples in row 3) and scenes (as shown in the room examples in row 1 and row 7).

We further compare our method with traditional approaches for converting low-resolution voxels to meshes. For the baselines, we first employ interpolation techniques such as nearest neighbor and trilinear interpolation, followed by the use of marching cubes~\cite{lorensen1998marching} to derive the meshes. Importantly, our approach is the first large-scale generative model to tackle this task. The quantitative and qualitative results of this comparison are presented in Table~\ref{tab:voxel_comp_tab} and Figure~\ref{fig:voxel_comp}. It is evident that, among the baseline methods, trilinear interpolation outperforms nearest neighbor, which is intuitively reasonable. Our method easily surpasses both of these traditional methods in terms of both IoU and LFD metrics.
%

% \lzz{may not need this paragraph.}%Additionally, it is noteworthy that voxel \(32^3\) achieves better results than voxel \(16^3\) in terms of the accuracy of the generated shapes compared to the ground truth. This outcome is expected, considering that the \(32^3\) voxel contains more information.

\subsection{3D Shape Completion}
\begin{figure}[t]
	\centering
	\includegraphics[width=1.0\columnwidth]{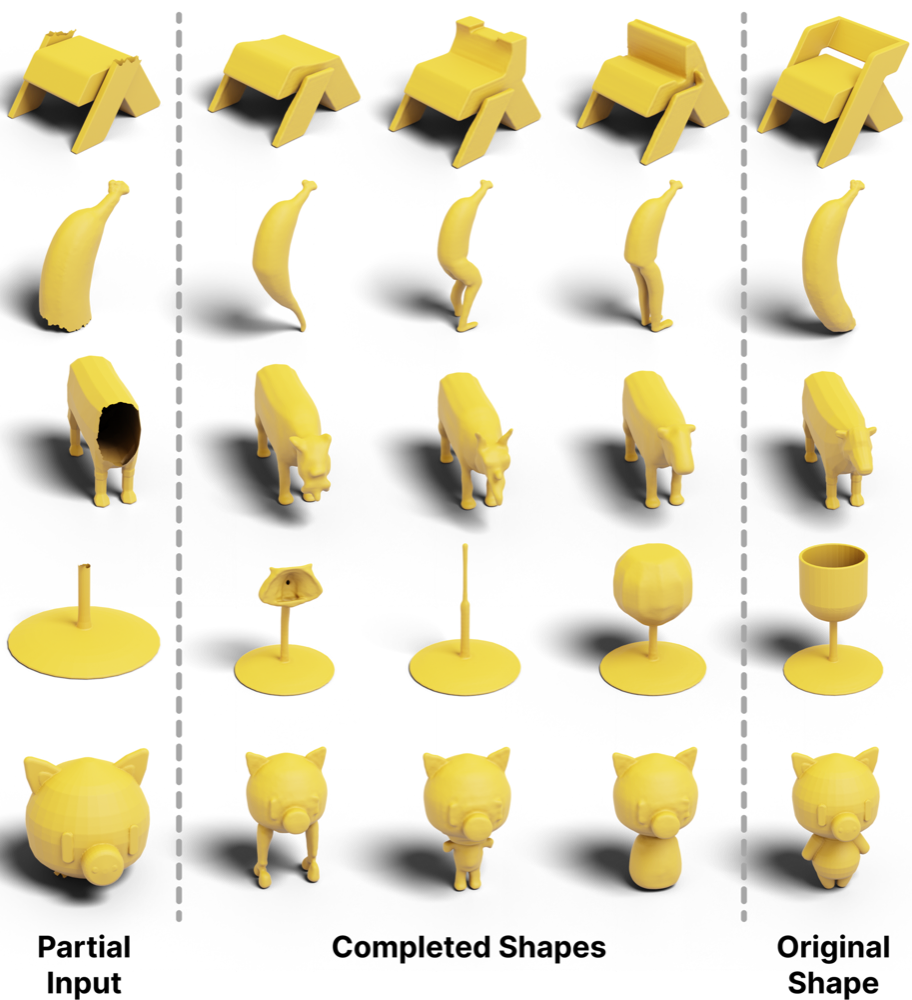}
	\caption{When provided with partial inputs (left), our unconditional generative model completes the missing regions in a coherent and semantically meaningful way. Additionally, it is capable of generating multiple variations of the completed shapes, many of which significantly differ from the original input (right), highlighting the diverse shape distribution learned.
}
\label{fig:completion_results}
\end{figure}

Furthermore, our trained unconditional generative model can be employed for completion tasks in a zero-shot fashion. In this context, the objective is to generate multiple variations of a partial input shape.
Given a shape, we first identify a region, consisting of a set of vertices on the mesh, that we aim to regenerate and subsequently discard it.
The remaining part of the mesh is retained as the partial input as shown in Figure~\ref{fig:completion_results} (leftmost column).
Subsequently, this partial shape is transformed into a TSDF, which is then converted into our diffusible wavelet representation with a grid structure.
Based on the earlier selected region for regeneration, we construct a binary mask that identifies missing regions to complete.
Utilizing the selection mask, we adopt the approach described in~\cite{lugmayr2022repaint} to regenerate the wavelet coefficients in these masked regions using our trained unconditional model.

Figure~\ref{fig:completion_results} shows visual examples of our completion approach.
It is evident that, in general, our unconditional model can generate semantically meaningful parts in the masked regions, as illustrated by the animal heads produced in the third row and the varied chair backs in the first row.
Additionally, these generated parts exhibit coherent connections to the given input of the partial shape, exemplified by the body region of the animal in the fifth row.
It is also noteworthy that our model can produce multiple variations for the same input partial shape, indicating the diverse distribution captured in our generative model.

\subsection{Ablation Studies}
\begin{table}
\centering
{\small
% \usepackage{multirow}
\caption{We quantitatively analyse the performance of our conditional generation models on different guidance weights.}
\resizebox{1.0\linewidth}{!}{
\begin{tabular}{c|c|ccccc}
\toprule
                              &                           & \multicolumn{5}{c}{Guidance Weight ($w$)}                                                                                                                                                               \\
\multirow{-2}{*}{Model}       & \multirow{-2}{*}{Metrics} & 1                                    & 2                                    & 3                                    & 4                                    & 5                                    \\
\midrule
                              & LFD $\downarrow$     & 4395.15  & \textbf{4071.33}                       & 4121.14  & 4192.30  & 4295.28  \\
\multirow{-2}{*}{Single-view} & IoU $\uparrow$     & 0.3706 & 0.4285                     & \textbf{0.4348} & 0.4289 & 0.4202 \\
\midrule
                              & LFD $\downarrow$     & 2378.48  & \textbf{2310.30}                       & 2413.18  & 2522.03  & 2639.69  \\
\multirow{-2}{*}{Multi-view}  & IoU $\uparrow$      & 0.6322 & \textbf{0.6595}                     & 0.6488 & 0.6317 & 0.6148 \\
\midrule
                              & LFD $\downarrow$     & 1701.17  & \textbf{1683.95}                      & 1769.93  & 1900.48   & 2029.59  \\
\multirow{-2}{*}{Voxels (32)} & IoU $\uparrow$      & 0.7636 & \textbf{0.7771}                     & 0.7659 & 0.7483  & 0.7323 \\
\midrule
                              & LFD $\downarrow$     & 2453.69      & \textbf{2347.04}  & 2426.40      & 2556.62      & 2724.72      \\
\multirow{-2}{*}{Voxels (16)} & IoU $\uparrow$      & 0.6424     & \textbf{0.6726} & 0.6614     & 0.6452     & 0.6289     \\
\midrule
                              & LFD $\downarrow$     & \textbf{1429.37}  & 1432.95                      & 1521.55  & 1658.03  & 1830.78  \\
\multirow{-2}{*}{Points}      & IoU $\uparrow$      & \textbf{0.8380} & 0.8379                         & 0.8207 & 0.8002 & 0.7781 \\
\bottomrule
\end{tabular}
}
\label{tab:scale_comp}
}
\end{table}

\begin{table}
\centering
{\small
% \usepackage{multirow}
\caption{We quantitatively evaluate the performances of generation models with different inference time steps.}
\resizebox{1.0\linewidth}{!}{
\begin{tabular}{c|c|cccc}
\toprule
                              &                           & \multicolumn{4}{c}{Inference Time step (t)}                                                                                                               \\
\multirow{-2}{*}{Model}       & \multirow{-2}{*}{Metrics} & 10                                   & 100                                  & 500                                  & 1000                                 \\
\midrule
                              & LFD $\downarrow$                       & 4312.23  & \textbf{4071.33}  & 4136.14    & 4113.10  \\
\multirow{-2}{*}{Single-view} & IoU $\uparrow$                       & \textbf{0.4477} & 0.4285 & 0.4186 & 0.4144 \\
\midrule
                              & LFD  $\downarrow$                      & \textbf{2217.25}   & 2310.30  & 2369.15  & 2394.17  \\
\multirow{-2}{*}{Multi-view}  & IoU  $\uparrow$                     & \textbf{0.6707} & 0.6595 & 0.6514 & 0.6445 \\
\midrule
                              & LFD $\downarrow$                       & \textbf{1580.98}  & 1683.95  & 1744.48  & 1763.91  \\
\multirow{-2}{*}{Voxels ($32^3$)} & IoU $\uparrow$                      & \textbf{0.7943} & 0.7771 & 0.7700  & 0.7667 \\
\midrule
                              & LFD  $\downarrow$                      & \textbf{2266.41}  & 2347.04                           & 2375.89                          & 2373.42                          \\
\multirow{-2}{*}{Voxels ($16^3$)} & IoU $\uparrow$                      & \textbf{0.6870} & 0.6726 & 0.6620 & 0.6616 \\
\midrule
                              & LFD $\downarrow$                       & \textbf{1368.90}  & 1429.37  & 1457.89  & 1468.91  \\
\multirow{-2}{*}{Point Cloud}      & IoU $\uparrow$                      & \textbf{0.8535} & 0.8380 & 0.8283 & 0.8287 \\
\midrule
Unconditional                 & FID $\downarrow$                       & 371.32  & 85.25  & 74.60   & \textbf{68.54} \\
\bottomrule
\end{tabular}
\label{tab:timestep_comp}
}
}
\end{table}

\begin{table}
\caption{
Employing Mean Squared Error (MSE) directly or separately for each subband resulted in worse performance compared to using just the coarse component~\cite{hui2022neural}, despite a higher theoretical representation capacity. In contrast, our subband adaptive training strategy led to significant improvements in both Light Field Distance (LFD) and Intersection Over Union (IoU) metrics.
}
\centering
{\fontsize{8}{9.6}\selectfont 
% \usepackage{multirow}

\begin{tabular}{c|cc}
\toprule
\multirow{2}{*}{Settings} & \multicolumn{2}{c}{Metrics} \\
                          & LFD $\downarrow$         & IoU $\uparrow$         \\
\midrule
Coarse component only~\cite{hui2022neural}                   & 2855.41  & 0.5919 \\
\midrule
Ours (MSE)                & 3191.49  & 0.5474  \\
Ours (subband-based MSE)         & 2824.28  & 0.5898 \\
\midrule
Ours                      & \textbf{2611.60}  & \textbf{0.6105} \\
\bottomrule
\end{tabular}
}
\label{tab:ablation}
\end{table}
% In the following, we perform various analyses on our generation framework.

\para{Classifier-free Guidance.}
As previously mentioned, we employ classifier-free guidance, as detailed in \cite{ho2022classifier}, to enhance the quality of conditioned samples. A crucial hyperparameter in this classifier-free guidance, during inference, is the scale parameter or guidance weight, denoted as \(w\). This parameter plays a key role in managing the trade-off between the generation's fidelity to the input conditions and the diversity of the generated output.

We experiment to explore the effect of the guidance weight parameter on the quality of samples generated by various conditional models. The guidance weight parameter was systematically adjusted in a linear progression from 1.0 to 5.0. It is important to note that, for efficient evaluation, an inference timestep of 100 was consistently employed across all experiments. The results of this study are presented in Table~\ref{tab:scale_comp}.

Empirically, we observe that a guidance weight of \(2.0\) is optimal for most conditional generation tasks. However, when the model is conditioned on point cloud data, a lower guidance weight of \(1.0\) yields better results. This contrasts with the text-to-image scenarios, which typically require a larger value for the guidance weight. We suspect this difference is attributable to the nature of the input conditions we use, such as images and point clouds, which contain more information and thus make it more challenging to generate diverse samples compared to text-based inputs.
Note that we adopt these identified optimal values as fixed hyperparameters for all subsequent inferences in the remainder of our experiments, as well as for the generation of qualitative results.

\para{Inference Time Step Analysis.}
Furthermore, we also provide a detailed analysis of the inference timesteps for both our conditional and unconditional models. Specifically, we evaluate the generation models under the same settings as above but with varying timesteps, namely 10, 100, 500, and 1000.

Table~\ref{tab:timestep_comp} presents the quantitative results for our different generative models using various time steps during inference.
Specifically, we empirically find that a small time step (10) suffices for conditions with minimal ambiguity, such as multi-view images, voxels, and point clouds. 
As ambiguity rises, the required number of time steps to achieve satisfactory sample quality also increases. 
For the unconditional model, which has no condition, the optimal time step is the maximum one (1000). 
Similarly to the guidance weight, we consider the optimal time step as a hyper-parameter, which is utilized in all experiments.

\para{Ablation of Adaptive Training Strategy.}
In this experiment, we use our multi-view model and initially compare our proposed representation with the generation results obtained using only the coarse component of our representation. This approach was adopted as the generation target in~\cite{hui2022neural}. This is shown in the first row of Table~\ref{tab:ablation}. 
Furthermore, to illustrate the effectiveness of our adopted subband adaptive training strategy, we also compare it with two other baseline training objectives.
% Moreover, to illustrate the effectiveness of our adopted subband adpative training strategy, we additionally compare with two baseline training objective.
%
First, we apply a direct MSE (Mean Squared Error) loss uniformly across all our coefficients on our diffusible representation. This approach is equivalent to the objective of the classical diffusion model.
Second, we contrast our strategy with a method that computes three separate MSE (Mean Squared Error) losses for $C_0$, $D_0$, and $D_1$, and then sums them directly.
It is important to note that separate models are necessary for each of these experiments, leading to increased computational resource costs. Additionally, convergence can take up to a few weeks, so we decided to implement an early stop at 750k iterations and compare the results at this stage.

From Table~\ref{tab:ablation}, it is observable that using solely the coarse component can still yield plausible results. This finding is in alignment with the discovery reported in~\cite{hui2022neural}.
However, naively applying MSE loss uniformly across all coefficients results in a performance drop, despite the upper bound of the representation capacity being significantly higher.
We observe that the initial attempt to separate the loss computation improves the performance of the trained generator. However, the quality of the results remains similar to those generated using only the coarse component.
By adopting our proposed subband adaptive training scheme, we achieve a significant improvement over the approach in~\cite{hui2022neural}.
% \lzz{Do we need any ablation on the model design for fidelity and efficient training？like the representation. Inference steps and CFG coefficient may be not crucial. Also, do we need any comparison with existing works like Wavelet and other 3D representations on fidelity and efficiency? }
% \edward{I think we have compared with wavelet in Table~\ref{tab:rep_compare} and Table~\ref{tab:ablation}. For the other representations, they are usually not scalable due to additional training.}

\section{Limitations and Future Works}
Our approach exhibits the following limitations:
(i) While our unconditional model is capable of generating a diverse variety of shapes from various sub-datasets, it can not ensure a balanced representation of objects across different categories during sampling.  Hence, the learned 3D shape distribution is inherently imbalanced, evident in the disproportionate representation of CAD models.
We can utilize a large zero-shot language model like ChatGPT for annotating object categories, enabling the application of diverse data augmentation methods to balance the training data according to these categories.
(ii) Our generation network, trained on a heterogeneous mix of datasets, does not utilize the category label as an extra condition.
Hence, our unconditional model may occasionally generate implausible shapes or introduce noise into the outputs.
Identifying and mitigating these anomalies represents a compelling direction for future research.
It is particularly intriguing to consider the development of data-driven metrics for assessing the visual plausibility of generated 3D shapes, especially in the context of the available large-scale 3D dataset.
(iii) At present, our primary focus lies in direct generation of 3D geometry. An interesting avenue for future exploration involves generating textures together on the geometry, with the aim of achieving this without relying on computationally-expensive optimizations.

\section{Conclusion}
In summary, this paper presents \nickname, a novel 3D generative framework trained on a vast dataset of over 10 millions publicly-available 3D shapes, capable of producing high-quality 3D shapes impressively within 2 seconds. Central to our approach is the introduction of a family of new techniques.
This includes the subband coefficient filtering scheme to help construct a compact, expressive, and efficient wavelet-tree representation that effectively encodes a $256^3$ SDF with minimal information loss. Then, we adeptly model the wavelet-tree representation by our diffusion-based generative model using our subband coefficient packing scheme, and further derive the subband adaptive training strategy to achieve model training that can effectively attends to both coarse and sparse detail coefficients. Besides, we also extend \nickname to take optional condition inputs of various modalities.

Our extensive experiments demonstrate the model's superiority in synthesizing high-quality 3D shapes across various challenging conditions, including single/multi-view images, point clouds, and low-resolution voxels, all while requiring minimal resource demands during the training. Remarkably, our model not only outperforms existing baselines quantitatively but also demonstrates zero-shot applications such as partial shape completion. We believe our work will pave the way for future research in other 3D representations to enable large-scale 3D model training.

{
    \small
    \bibliographystyle{ieeenat_fullname}
    \bibliography{main}

\begin{thebibliography}{110}
\providecommand{\natexlab}[1]{#1}
\providecommand{\url}[1]{\texttt{#1}}
\expandafter\ifx\csname urlstyle\endcsname\relax
  \providecommand{\doi}[1]{doi: #1}\else
  \providecommand{\doi}{doi: \begingroup \urlstyle{rm}\Url}\fi

\bibitem[Achlioptas et~al.(2018)Achlioptas, Diamanti, Mitliagkas, and Guibas]{achlioptas2018learning}
Panos Achlioptas, Olga Diamanti, Ioannis Mitliagkas, and Leonidas Guibas.
\newblock Learning representations and generative models for 3d point clouds.
\newblock In \emph{International conference on machine learning}, pages 40--49. PMLR, 2018.

\bibitem[Chan et~al.(2022)Chan, Lin, Chan, Nagano, Pan, De~Mello, Gallo, Guibas, Tremblay, Khamis, et~al.]{chan2022efficient}
Eric~R Chan, Connor~Z Lin, Matthew~A Chan, Koki Nagano, Boxiao Pan, Shalini De~Mello, Orazio Gallo, Leonidas~J Guibas, Jonathan Tremblay, Sameh Khamis, et~al.
\newblock Efficient geometry-aware 3d generative adversarial networks.
\newblock In \emph{Proceedings of the IEEE/CVF Conference on Computer Vision and Pattern Recognition}, pages 16123--16133, 2022.

\bibitem[Chang et~al.(2015)Chang, Funkhouser, Guibas, Hanrahan, Huang, Li, Savarese, Savva, Song, Su, et~al.]{chang2015shapenet}
Angel~X Chang, Thomas Funkhouser, Leonidas Guibas, Pat Hanrahan, Qixing Huang, Zimo Li, Silvio Savarese, Manolis Savva, Shuran Song, Hao Su, et~al.
\newblock Shapenet: An information-rich 3d model repository.
\newblock \emph{arXiv preprint arXiv:1512.03012}, 2015.

\bibitem[Chen et~al.(2003)Chen, Tian, Shen, and Ouhyoung]{chen2003visual}
Ding-Yun Chen, Xiao-Pei Tian, Yu-Te Shen, and Ming Ouhyoung.
\newblock On visual similarity based 3d model retrieval.
\newblock In \emph{Computer graphics forum}, pages 223--232. Wiley Online Library, 2003.

\bibitem[Chen et~al.(2023)Chen, Chen, Zhou, and Zhang]{chen2023shaddr}
Qimin Chen, Zhiqin Chen, Hang Zhou, and Hao Zhang.
\newblock Shaddr: Real-time example-based geometry and texture generation via 3d shape detailization and differentiable rendering.
\newblock \emph{arXiv preprint arXiv:2306.04889}, 2023.

\bibitem[Chen and Zhang(2019)]{chen2019learning}
Zhiqin Chen and Hao Zhang.
\newblock Learning implicit fields for generative shape modeling.
\newblock In \emph{Proceedings of the IEEE/CVF Conference on Computer Vision and Pattern Recognition}, pages 5939--5948, 2019.

\bibitem[Chen et~al.(2021)Chen, Kim, Fisher, Aigerman, Zhang, and Chaudhuri]{chen2021decor}
Zhiqin Chen, Vladimir~G Kim, Matthew Fisher, Noam Aigerman, Hao Zhang, and Siddhartha Chaudhuri.
\newblock Decor-gan: 3d shape detailization by conditional refinement.
\newblock In \emph{Proceedings of the IEEE/CVF conference on computer vision and pattern recognition}, pages 15740--15749, 2021.

\bibitem[Cheng et~al.(2022)Cheng, Li, Liu, Sun, and Yang]{cheng2022autoregressive}
An-Chieh Cheng, Xueting Li, Sifei Liu, Min Sun, and Ming-Hsuan Yang.
\newblock Autoregressive 3d shape generation via canonical mapping.
\newblock In \emph{European Conference on Computer Vision}, pages 89--104. Springer, 2022.

\bibitem[Cheng et~al.(2023)Cheng, Lee, Tulyakov, Schwing, and Gui]{cheng2023sdfusion}
Yen-Chi Cheng, Hsin-Ying Lee, Sergey Tulyakov, Alexander~G Schwing, and Liang-Yan Gui.
\newblock Sdfusion: Multimodal 3d shape completion, reconstruction, and generation.
\newblock In \emph{Proceedings of the IEEE/CVF Conference on Computer Vision and Pattern Recognition}, pages 4456--4465, 2023.

\bibitem[Chou et~al.(2023)Chou, Bahat, and Heide]{chou2023diffusion}
Gene Chou, Yuval Bahat, and Felix Heide.
\newblock Diffusion-sdf: Conditional generative modeling of signed distance functions.
\newblock In \emph{Proceedings of the IEEE/CVF International Conference on Computer Vision}, pages 2262--2272, 2023.

\bibitem[Collins et~al.(2022)Collins, Goel, Deng, Luthra, Xu, Gundogdu, Zhang, Vicente, Dideriksen, Arora, et~al.]{collins2022abo}
Jasmine Collins, Shubham Goel, Kenan Deng, Achleshwar Luthra, Leon Xu, Erhan Gundogdu, Xi Zhang, Tomas F~Yago Vicente, Thomas Dideriksen, Himanshu Arora, et~al.
\newblock Abo: Dataset and benchmarks for real-world 3d object understanding.
\newblock In \emph{Proceedings of the IEEE/CVF Conference on Computer Vision and Pattern Recognition}, pages 21126--21136, 2022.

\bibitem[Deitke et~al.(2023)Deitke, Schwenk, Salvador, Weihs, Michel, VanderBilt, Schmidt, Ehsani, Kembhavi, and Farhadi]{deitke2023objaverse}
Matt Deitke, Dustin Schwenk, Jordi Salvador, Luca Weihs, Oscar Michel, Eli VanderBilt, Ludwig Schmidt, Kiana Ehsani, Aniruddha Kembhavi, and Ali Farhadi.
\newblock Objaverse: A universe of annotated 3d objects.
\newblock In \emph{Proceedings of the IEEE/CVF Conference on Computer Vision and Pattern Recognition}, pages 13142--13153, 2023.

\bibitem[Deng et~al.(2023)Deng, Jiang, Qi, Yan, Zhou, Guibas, Anguelov, et~al.]{deng2023nerdi}
Congyue Deng, Chiyu Jiang, Charles~R Qi, Xinchen Yan, Yin Zhou, Leonidas Guibas, Dragomir Anguelov, et~al.
\newblock Nerdi: Single-view nerf synthesis with language-guided diffusion as general image priors.
\newblock In \emph{Proceedings of the IEEE/CVF Conference on Computer Vision and Pattern Recognition}, pages 20637--20647, 2023.

\bibitem[Dhariwal and Nichol(2021)]{dhariwal2021diffusion}
Prafulla Dhariwal and Alexander Nichol.
\newblock Diffusion models beat gans on image synthesis.
\newblock \emph{NeurIPS}, 34:\penalty0 8780--8794, 2021.

\bibitem[Downs et~al.(2022)Downs, Francis, Koenig, Kinman, Hickman, Reymann, McHugh, and Vanhoucke]{downs2022google}
Laura Downs, Anthony Francis, Nate Koenig, Brandon Kinman, Ryan Hickman, Krista Reymann, Thomas~B McHugh, and Vincent Vanhoucke.
\newblock Google scanned objects: A high-quality dataset of 3d scanned household items.
\newblock In \emph{2022 International Conference on Robotics and Automation (ICRA)}, pages 2553--2560. IEEE, 2022.

\bibitem[Fu et~al.(2021)Fu, Jia, Gao, Gong, Zhao, Maybank, and Tao]{fu20213d}
Huan Fu, Rongfei Jia, Lin Gao, Mingming Gong, Binqiang Zhao, Steve Maybank, and Dacheng Tao.
\newblock 3d-future: 3d furniture shape with texture.
\newblock \emph{International Journal of Computer Vision}, 129:\penalty0 3313--3337, 2021.

\bibitem[Gao et~al.(2022{\natexlab{a}})Gao, Yu, Sheng, Song, and Xu]{gao2022sketchsampler}
Chenjian Gao, Qian Yu, Lu Sheng, Yi-Zhe Song, and Dong Xu.
\newblock Sketchsampler: Sketch-based 3d reconstruction via view-dependent depth sampling.
\newblock In \emph{Computer Vision--ECCV 2022: 17th European Conference, Tel Aviv, Israel, October 23--27, 2022, Proceedings, Part I}, pages 464--479. Springer, 2022{\natexlab{a}}.

\bibitem[Gao et~al.(2022{\natexlab{b}})Gao, Shen, Wang, Chen, Yin, Li, Litany, Gojcic, and Fidler]{gao2022get3d}
Jun Gao, Tianchang Shen, Zian Wang, Wenzheng Chen, Kangxue Yin, Daiqing Li, Or Litany, Zan Gojcic, and Sanja Fidler.
\newblock Get3d: A generative model of high quality 3d textured shapes learned from images.
\newblock \emph{Advances In Neural Information Processing Systems}, 35:\penalty0 31841--31854, 2022{\natexlab{b}}.

\bibitem[Goodfellow et~al.(2014)Goodfellow, Pouget-Abadie, Mirza, Xu, Warde-Farley, Ozair, Courville, and Bengio]{goodfellow2014generative}
Ian Goodfellow, Jean Pouget-Abadie, Mehdi Mirza, Bing Xu, David Warde-Farley, Sherjil Ozair, Aaron Courville, and Yoshua Bengio.
\newblock Generative adversarial nets.
\newblock \emph{Advances in neural information processing systems}, 27, 2014.

\bibitem[Guillard et~al.(2021)Guillard, Remelli, Yvernay, and Fua]{guillard2021sketch2mesh}
Benoit Guillard, Edoardo Remelli, Pierre Yvernay, and Pascal Fua.
\newblock Sketch2mesh: Reconstructing and editing 3d shapes from sketches.
\newblock In \emph{Proceedings of the IEEE/CVF International Conference on Computer Vision}, pages 13023--13032, 2021.

\bibitem[Hanocka et~al.(2019)Hanocka, Hertz, Fish, Giryes, Fleishman, and Cohen-Or]{hanocka2019meshcnn}
Rana Hanocka, Amir Hertz, Noa Fish, Raja Giryes, Shachar Fleishman, and Daniel Cohen-Or.
\newblock Meshcnn: a network with an edge.
\newblock \emph{ACM Transactions on Graphics (ToG)}, 38\penalty0 (4):\penalty0 1--12, 2019.

\bibitem[Heusel et~al.(2017)Heusel, Ramsauer, Unterthiner, Nessler, and Hochreiter]{heusel2017gans}
Martin Heusel, Hubert Ramsauer, Thomas Unterthiner, Bernhard Nessler, and Sepp Hochreiter.
\newblock Gans trained by a two time-scale update rule converge to a local nash equilibrium.
\newblock \emph{Advances in neural information processing systems}, 30, 2017.

\bibitem[Ho and Salimans(2022)]{ho2022classifier}
Jonathan Ho and Tim Salimans.
\newblock Classifier-free diffusion guidance.
\newblock \emph{arXiv preprint arXiv:2207.12598}, 2022.

\bibitem[Ho et~al.(2020)Ho, Jain, and Abbeel]{ho2020denoising}
Jonathan Ho, Ajay Jain, and Pieter Abbeel.
\newblock Denoising diffusion probabilistic models.
\newblock \emph{Advances in neural information processing systems}, 33:\penalty0 6840--6851, 2020.

\bibitem[Hong et~al.(2023)Hong, Zhang, Gu, Bi, Zhou, Liu, Liu, Sunkavalli, Bui, and Tan]{hong2023lrm}
Yicong Hong, Kai Zhang, Jiuxiang Gu, Sai Bi, Yang Zhou, Difan Liu, Feng Liu, Kalyan Sunkavalli, Trung Bui, and Hao Tan.
\newblock Lrm: Large reconstruction model for single image to 3d.
\newblock \emph{arXiv preprint arXiv:2311.04400}, 2023.

\bibitem[Hoogeboom et~al.(2023)Hoogeboom, Heek, and Salimans]{hoogeboom2023simple}
Emiel Hoogeboom, Jonathan Heek, and Tim Salimans.
\newblock simple diffusion: End-to-end diffusion for high resolution images.
\newblock \emph{arXiv preprint arXiv:2301.11093}, 2023.

\bibitem[Huang et~al.(2023)Huang, Zhou, Yan, and Lin]{huang2023scalelong}
Zhongzhan Huang, Pan Zhou, Shuicheng Yan, and Liang Lin.
\newblock Scalelong: Towards more stable training of diffusion model via scaling network long skip connection.
\newblock \emph{arXiv preprint arXiv:2310.13545}, 2023.

\bibitem[Hui et~al.(2022)Hui, Li, Hu, and Fu]{hui2022neural}
Ka-Hei Hui, Ruihui Li, Jingyu Hu, and Chi-Wing Fu.
\newblock Neural wavelet-domain diffusion for {3D} shape generation.
\newblock In \emph{ACM SIGGRAPH Asia}, pages 1--9, 2022.

\bibitem[Ibing et~al.(2021)Ibing, Lim, and Kobbelt]{ibing20213d}
Moritz Ibing, Isaak Lim, and Leif Kobbelt.
\newblock 3d shape generation with grid-based implicit functions.
\newblock In \emph{Proceedings of the IEEE/CVF Conference on Computer Vision and Pattern Recognition}, pages 13559--13568, 2021.

\bibitem[Jain et~al.(2022)Jain, Mildenhall, Barron, Abbeel, and Poole]{jain2022zero}
Ajay Jain, Ben Mildenhall, Jonathan~T Barron, Pieter Abbeel, and Ben Poole.
\newblock Zero-shot text-guided object generation with dream fields.
\newblock In \emph{Proceedings of the IEEE/CVF Conference on Computer Vision and Pattern Recognition}, pages 867--876, 2022.

\bibitem[Jayaraman et~al.(2021)Jayaraman, Sanghi, Lambourne, Willis, Davies, Shayani, and Morris]{jayaraman2021uv}
Pradeep~Kumar Jayaraman, Aditya Sanghi, Joseph~G Lambourne, Karl~DD Willis, Thomas Davies, Hooman Shayani, and Nigel Morris.
\newblock Uv-net: Learning from boundary representations.
\newblock In \emph{Proceedings of the IEEE/CVF conference on computer vision and pattern recognition}, pages 11703--11712, 2021.

\bibitem[Jayaraman et~al.(2022)Jayaraman, Lambourne, Desai, Willis, Sanghi, and Morris]{jayaraman2022solidgen}
Pradeep~Kumar Jayaraman, Joseph~G Lambourne, Nishkrit Desai, Karl~DD Willis, Aditya Sanghi, and Nigel~JW Morris.
\newblock Solidgen: An autoregressive model for direct b-rep synthesis.
\newblock \emph{arXiv preprint arXiv:2203.13944}, 2022.

\bibitem[Jun and Nichol(2023)]{jun2023shap}
Heewoo Jun and Alex Nichol.
\newblock Shap-e: Generating conditional {3D} implicit functions.
\newblock \emph{arXiv preprint arXiv:2305.02463}, 2023.

\bibitem[Kingma and Ba(2014)]{kingma2014adam}
Diederik~P Kingma and Jimmy Ba.
\newblock Adam: A method for stochastic optimization.
\newblock \emph{arXiv preprint arXiv:1412.6980}, 2014.

\bibitem[Klokov et~al.(2020)Klokov, Boyer, and Verbeek]{klokov2020discrete}
Roman Klokov, Edmond Boyer, and Jakob Verbeek.
\newblock Discrete point flow networks for efficient point cloud generation.
\newblock In \emph{European Conference on Computer Vision}, pages 694--710. Springer, 2020.

\bibitem[Koch et~al.(2019)Koch, Matveev, Jiang, Williams, Artemov, Burnaev, Alexa, Zorin, and Panozzo]{koch2019abc}
Sebastian Koch, Albert Matveev, Zhongshi Jiang, Francis Williams, Alexey Artemov, Evgeny Burnaev, Marc Alexa, Denis Zorin, and Daniele Panozzo.
\newblock Abc: A big cad model dataset for geometric deep learning.
\newblock In \emph{Proceedings of the IEEE/CVF conference on computer vision and pattern recognition}, pages 9601--9611, 2019.

\bibitem[Kong et~al.(2022)Kong, Wang, and Qi]{kong2022diffusion}
Di Kong, Qiang Wang, and Yonggang Qi.
\newblock A diffusion-refinement model for sketch-to-point modeling.
\newblock In \emph{Proceedings of the Asian Conference on Computer Vision}, pages 1522--1538, 2022.

\bibitem[Lambourne et~al.(2021)Lambourne, Willis, Jayaraman, Sanghi, Meltzer, and Shayani]{lambourne2021brepnet}
Joseph~G Lambourne, Karl~DD Willis, Pradeep~Kumar Jayaraman, Aditya Sanghi, Peter Meltzer, and Hooman Shayani.
\newblock Brepnet: A topological message passing system for solid models.
\newblock In \emph{Proceedings of the IEEE/CVF conference on computer vision and pattern recognition}, pages 12773--12782, 2021.

\bibitem[Lee et~al.(2019)Lee, Lee, Kim, Kosiorek, Choi, and Teh]{lee2019set}
Juho Lee, Yoonho Lee, Jungtaek Kim, Adam Kosiorek, Seungjin Choi, and Yee~Whye Teh.
\newblock Set transformer: A framework for attention-based permutation-invariant neural networks.
\newblock In \emph{International conference on machine learning}, pages 3744--3753. PMLR, 2019.

\bibitem[Li et~al.(2023{\natexlab{a}})Li, Tan, Zhang, Xu, Luan, Xu, Hong, Sunkavalli, Shakhnarovich, and Bi]{li2023instant3d}
Jiahao Li, Hao Tan, Kai Zhang, Zexiang Xu, Fujun Luan, Yinghao Xu, Yicong Hong, Kalyan Sunkavalli, Greg Shakhnarovich, and Sai Bi.
\newblock Instant3d: Fast text-to-3d with sparse-view generation and large reconstruction model.
\newblock \emph{arXiv preprint arXiv:2311.06214}, 2023{\natexlab{a}}.

\bibitem[Li et~al.(2023{\natexlab{b}})Li, Duan, Zhou, and Lu]{li2023diffusion}
Muheng Li, Yueqi Duan, Jie Zhou, and Jiwen Lu.
\newblock Diffusion-sdf: Text-to-shape via voxelized diffusion.
\newblock In \emph{Proceedings of the IEEE/CVF Conference on Computer Vision and Pattern Recognition}, pages 12642--12651, 2023{\natexlab{b}}.

\bibitem[Li et~al.(2021)Li, Takehara, Taketomi, Zheng, and Nie{\ss}ner]{li20214dcomplete}
Yang Li, Hikari Takehara, Takafumi Taketomi, Bo Zheng, and Matthias Nie{\ss}ner.
\newblock 4dcomplete: Non-rigid motion estimation beyond the observable surface.
\newblock In \emph{Proceedings of the IEEE/CVF International Conference on Computer Vision}, pages 12706--12716, 2021.

\bibitem[Liu et~al.(2023{\natexlab{a}})Liu, Xu, Jin, Chen, Xu, Su, et~al.]{liu2023one}
Minghua Liu, Chao Xu, Haian Jin, Linghao Chen, Zexiang Xu, Hao Su, et~al.
\newblock One-2-3-45: Any single image to 3d mesh in 45 seconds without per-shape optimization.
\newblock \emph{arXiv preprint arXiv:2306.16928}, 2023{\natexlab{a}}.

\bibitem[Liu et~al.(2023{\natexlab{b}})Liu, Wu, Van~Hoorick, Tokmakov, Zakharov, and Vondrick]{liu2023zero}
Ruoshi Liu, Rundi Wu, Basile Van~Hoorick, Pavel Tokmakov, Sergey Zakharov, and Carl Vondrick.
\newblock Zero-1-to-3: Zero-shot one image to 3d object.
\newblock In \emph{Proceedings of the IEEE/CVF International Conference on Computer Vision}, pages 9298--9309, 2023{\natexlab{b}}.

\bibitem[Liu et~al.(2021)Liu, Han, Liu, and Zwicker]{liu2021fine}
Xinhai Liu, Zhizhong Han, Yu-Shen Liu, and Matthias Zwicker.
\newblock Fine-grained 3d shape classification with hierarchical part-view attention.
\newblock \emph{IEEE Transactions on Image Processing}, 30:\penalty0 1744--1758, 2021.

\bibitem[Liu et~al.(2022)Liu, Dai, Li, Qi, and Fu]{liu2022iss}
Zhengzhe Liu, Peng Dai, Ruihui Li, Xiaojuan Qi, and Chi-Wing Fu.
\newblock Iss: Image as stetting stone for text-guided 3d shape generation.
\newblock \emph{arXiv preprint arXiv:2209.04145}, 2022.

\bibitem[Liu et~al.(2023{\natexlab{c}})Liu, Hu, Hui, Qi, Cohen-Or, and Fu]{liu2023exim}
Zhengzhe Liu, Jingyu Hu, Ka-Hei Hui, Xiaojuan Qi, Daniel Cohen-Or, and Chi-Wing Fu.
\newblock Exim: A hybrid explicit-implicit representation for text-guided 3d shape generation.
\newblock \emph{ACM Transactions on Graphics (TOG)}, 42\penalty0 (6):\penalty0 1--12, 2023{\natexlab{c}}.

\bibitem[Loper et~al.(2015)Loper, Mahmood, Romero, Pons-Moll, and Black]{loper2015smpl}
Matthew Loper, Naureen Mahmood, Javier Romero, Gerard Pons-Moll, and Michael~J. Black.
\newblock {SMPL}: A skinned multi-person linear model.
\newblock \emph{ACM Trans. Graphics (Proc. SIGGRAPH Asia)}, 34\penalty0 (6):\penalty0 248:1--248:16, 2015.

\bibitem[Lorensen and Cline(1998)]{lorensen1998marching}
William~E Lorensen and Harvey~E Cline.
\newblock Marching cubes: A high resolution 3d surface construction algorithm.
\newblock In \emph{Seminal graphics: pioneering efforts that shaped the field}, pages 347--353. 1998.

\bibitem[Lugmayr et~al.(2022)Lugmayr, Danelljan, Romero, Yu, Timofte, and Van~Gool]{lugmayr2022repaint}
Andreas Lugmayr, Martin Danelljan, Andres Romero, Fisher Yu, Radu Timofte, and Luc Van~Gool.
\newblock Repaint: Inpainting using denoising diffusion probabilistic models.
\newblock In \emph{Proceedings of the IEEE/CVF Conference on Computer Vision and Pattern Recognition}, pages 11461--11471, 2022.

\bibitem[Lun et~al.(2017)Lun, Gadelha, Kalogerakis, Maji, and Wang]{lun20173d}
Zhaoliang Lun, Matheus Gadelha, Evangelos Kalogerakis, Subhransu Maji, and Rui Wang.
\newblock 3d shape reconstruction from sketches via multi-view convolutional networks.
\newblock In \emph{2017 International Conference on 3D Vision (3DV)}, pages 67--77. IEEE, 2017.

\bibitem[Luo and Hu(2021)]{luo2021diffusion}
Shitong Luo and Wei Hu.
\newblock Diffusion probabilistic models for 3d point cloud generation.
\newblock In \emph{Proceedings of the IEEE/CVF Conference on Computer Vision and Pattern Recognition}, pages 2837--2845, 2021.

\bibitem[Masci et~al.(2015)Masci, Boscaini, Bronstein, and Vandergheynst]{masci2015geodesic}
Jonathan Masci, Davide Boscaini, Michael Bronstein, and Pierre Vandergheynst.
\newblock Geodesic convolutional neural networks on riemannian manifolds.
\newblock In \emph{Proceedings of the IEEE international conference on computer vision workshops}, pages 37--45, 2015.

\bibitem[Maturana and Scherer(2015)]{maturana2015voxnet}
Daniel Maturana and Sebastian Scherer.
\newblock Voxnet: A 3d convolutional neural network for real-time object recognition.
\newblock In \emph{2015 IEEE/RSJ international conference on intelligent robots and systems (IROS)}, pages 922--928. IEEE, 2015.

\bibitem[Melas-Kyriazi et~al.(2023)Melas-Kyriazi, Laina, Rupprecht, and Vedaldi]{melas2023realfusion}
Luke Melas-Kyriazi, Iro Laina, Christian Rupprecht, and Andrea Vedaldi.
\newblock Realfusion: 360deg reconstruction of any object from a single image.
\newblock In \emph{Proceedings of the IEEE/CVF Conference on Computer Vision and Pattern Recognition}, pages 8446--8455, 2023.

\bibitem[Mescheder et~al.(2019)Mescheder, Oechsle, Niemeyer, Nowozin, and Geiger]{mescheder2019occupancy}
Lars Mescheder, Michael Oechsle, Michael Niemeyer, Sebastian Nowozin, and Andreas Geiger.
\newblock Occupancy networks: Learning 3d reconstruction in function space.
\newblock In \emph{Proceedings of the IEEE/CVF conference on computer vision and pattern recognition}, pages 4460--4470, 2019.

\bibitem[Michel et~al.(2022)Michel, Bar-On, Liu, Benaim, and Hanocka]{michel2022text2mesh}
Oscar Michel, Roi Bar-On, Richard Liu, Sagie Benaim, and Rana Hanocka.
\newblock Text2mesh: Text-driven neural stylization for meshes.
\newblock In \emph{Proceedings of the IEEE/CVF Conference on Computer Vision and Pattern Recognition}, pages 13492--13502, 2022.

\bibitem[Mikaeili et~al.(2023)Mikaeili, Perel, Safaee, Cohen-Or, and Mahdavi-Amiri]{mikaeili2023sked}
Aryan Mikaeili, Or Perel, Mehdi Safaee, Daniel Cohen-Or, and Ali Mahdavi-Amiri.
\newblock Sked: Sketch-guided text-based 3d editing.
\newblock In \emph{Proceedings of the IEEE/CVF International Conference on Computer Vision}, pages 14607--14619, 2023.

\bibitem[Mittal et~al.(2022)Mittal, Cheng, Singh, and Tulsiani]{mittal2022autosdf}
Paritosh Mittal, Yen-Chi Cheng, Maneesh Singh, and Shubham Tulsiani.
\newblock Autosdf: Shape priors for 3d completion, reconstruction and generation.
\newblock In \emph{Proceedings of the IEEE/CVF Conference on Computer Vision and Pattern Recognition}, pages 306--315, 2022.

\bibitem[Mo et~al.(2019)Mo, Guerrero, Yi, Su, Wonka, Mitra, and Guibas]{mo2019structurenet}
Kaichun Mo, Paul Guerrero, Li Yi, Hao Su, Peter Wonka, Niloy Mitra, and Leonidas~J Guibas.
\newblock Structurenet: Hierarchical graph networks for 3d shape generation.
\newblock \emph{arXiv preprint arXiv:1908.00575}, 2019.

\bibitem[Nash et~al.(2020)Nash, Ganin, Eslami, and Battaglia]{nash2020polygen}
Charlie Nash, Yaroslav Ganin, SM~Ali Eslami, and Peter Battaglia.
\newblock Polygen: An autoregressive generative model of 3d meshes.
\newblock In \emph{International conference on machine learning}, pages 7220--7229. PMLR, 2020.

\bibitem[Nichol et~al.(2022)Nichol, Jun, Dhariwal, Mishkin, and Chen]{nichol2022point}
Alex Nichol, Heewoo Jun, Prafulla Dhariwal, Pamela Mishkin, and Mark Chen.
\newblock Point-e: A system for generating 3d point clouds from complex prompts.
\newblock \emph{arXiv preprint arXiv:2212.08751}, 2022.

\bibitem[Park et~al.(2019)Park, Florence, Straub, Newcombe, and Lovegrove]{park2019deepsdf}
Jeong~Joon Park, Peter Florence, Julian Straub, Richard Newcombe, and Steven Lovegrove.
\newblock Deepsdf: Learning continuous signed distance functions for shape representation.
\newblock In \emph{Proceedings of the IEEE/CVF conference on computer vision and pattern recognition}, pages 165--174, 2019.

\bibitem[Peng et~al.(2020)Peng, Niemeyer, Mescheder, Pollefeys, and Geiger]{peng2020convolutional}
Songyou Peng, Michael Niemeyer, Lars Mescheder, Marc Pollefeys, and Andreas Geiger.
\newblock Convolutional occupancy networks.
\newblock In \emph{Computer Vision--ECCV 2020: 16th European Conference, Glasgow, UK, August 23--28, 2020, Proceedings, Part III 16}, pages 523--540. Springer, 2020.

\bibitem[Poole et~al.(2022)Poole, Jain, Barron, and Mildenhall]{poole2022dreamfusion}
Ben Poole, Ajay Jain, Jonathan~T Barron, and Ben Mildenhall.
\newblock Dreamfusion: Text-to-3d using 2d diffusion.
\newblock \emph{arXiv preprint arXiv:2209.14988}, 2022.

\bibitem[Qi et~al.(2016)Qi, Su, Nie{\ss}ner, Dai, Yan, and Guibas]{qi2016volumetric}
Charles~R Qi, Hao Su, Matthias Nie{\ss}ner, Angela Dai, Mengyuan Yan, and Leonidas~J Guibas.
\newblock Volumetric and multi-view cnns for object classification on 3d data.
\newblock In \emph{Proceedings of the IEEE conference on computer vision and pattern recognition}, pages 5648--5656, 2016.

\bibitem[Qi et~al.(2017{\natexlab{a}})Qi, Su, Mo, and Guibas]{qi2017pointnet}
Charles~R Qi, Hao Su, Kaichun Mo, and Leonidas~J Guibas.
\newblock Pointnet: Deep learning on point sets for 3d classification and segmentation.
\newblock In \emph{Proceedings of the IEEE conference on computer vision and pattern recognition}, pages 652--660, 2017{\natexlab{a}}.

\bibitem[Qi et~al.(2017{\natexlab{b}})Qi, Yi, Su, and Guibas]{qi2017pointnet++}
Charles~Ruizhongtai Qi, Li Yi, Hao Su, and Leonidas~J Guibas.
\newblock Pointnet++: Deep hierarchical feature learning on point sets in a metric space.
\newblock \emph{Advances in neural information processing systems}, 30, 2017{\natexlab{b}}.

\bibitem[Qian et~al.(2023)Qian, Mai, Hamdi, Ren, Siarohin, Li, Lee, Skorokhodov, Wonka, Tulyakov, et~al.]{qian2023magic123}
Guocheng Qian, Jinjie Mai, Abdullah Hamdi, Jian Ren, Aliaksandr Siarohin, Bing Li, Hsin-Ying Lee, Ivan Skorokhodov, Peter Wonka, Sergey Tulyakov, et~al.
\newblock Magic123: One image to high-quality 3d object generation using both 2d and 3d diffusion priors.
\newblock \emph{arXiv preprint arXiv:2306.17843}, 2023.

\bibitem[Radford et~al.(2021)Radford, Kim, Hallacy, Ramesh, Goh, Agarwal, Sastry, Askell, Mishkin, Clark, et~al.]{radford2021learning}
Alec Radford, Jong~Wook Kim, Chris Hallacy, Aditya Ramesh, Gabriel Goh, Sandhini Agarwal, Girish Sastry, Amanda Askell, Pamela Mishkin, Jack Clark, et~al.
\newblock Learning transferable visual models from natural language supervision.
\newblock In \emph{International conference on machine learning}, pages 8748--8763. PMLR, 2021.

\bibitem[Raistrick et~al.(2023)Raistrick, Lipson, Ma, Mei, Wang, Zuo, Kayan, Wen, Han, Wang, et~al.]{raistrick2023infinite}
Alexander Raistrick, Lahav Lipson, Zeyu Ma, Lingjie Mei, Mingzhe Wang, Yiming Zuo, Karhan Kayan, Hongyu Wen, Beining Han, Yihan Wang, et~al.
\newblock Infinite photorealistic worlds using procedural generation.
\newblock In \emph{Proceedings of the IEEE/CVF Conference on Computer Vision and Pattern Recognition}, pages 12630--12641, 2023.

\bibitem[Ramesh et~al.(2021)Ramesh, Pavlov, Goh, Gray, Voss, Radford, Chen, and Sutskever]{ramesh2021zero}
Aditya Ramesh, Mikhail Pavlov, Gabriel Goh, Scott Gray, Chelsea Voss, Alec Radford, Mark Chen, and Ilya Sutskever.
\newblock Zero-shot text-to-image generation.
\newblock In \emph{International Conference on Machine Learning}, pages 8821--8831. PMLR, 2021.

\bibitem[Ranjan et~al.(2018)Ranjan, Bolkart, Sanyal, and Black]{ranjan2018coma}
Anurag Ranjan, Timo Bolkart, Soubhik Sanyal, and Michael~J. Black.
\newblock Generating {3D} faces using convolutional mesh autoencoders.
\newblock In \emph{European Conference on Computer Vision (ECCV)}, pages 725--741, 2018.

\bibitem[Rombach et~al.(2022)Rombach, Blattmann, Lorenz, Esser, and Ommer]{rombach2022high}
Robin Rombach, Andreas Blattmann, Dominik Lorenz, Patrick Esser, and Bj{\"o}rn Ommer.
\newblock High-resolution image synthesis with latent diffusion models.
\newblock In \emph{Proceedings of the IEEE/CVF conference on computer vision and pattern recognition}, pages 10684--10695, 2022.

\bibitem[Saharia et~al.(2022)Saharia, Chan, Saxena, Li, Whang, Denton, Ghasemipour, Gontijo~Lopes, Karagol~Ayan, Salimans, et~al.]{saharia2022photorealistic}
Chitwan Saharia, William Chan, Saurabh Saxena, Lala Li, Jay Whang, Emily~L Denton, Kamyar Ghasemipour, Raphael Gontijo~Lopes, Burcu Karagol~Ayan, Tim Salimans, et~al.
\newblock Photorealistic text-to-image diffusion models with deep language understanding.
\newblock \emph{Advances in Neural Information Processing Systems}, 35:\penalty0 36479--36494, 2022.

\bibitem[Sanghi et~al.(2022)Sanghi, Chu, Lambourne, Wang, Cheng, Fumero, and Malekshan]{sanghi2022clip}
Aditya Sanghi, Hang Chu, Joseph~G. Lambourne, Ye Wang, Chin-Yi Cheng, Marco Fumero, and Kamal~Rahimi Malekshan.
\newblock {CLIP-Forge}: Towards zero-shot text-to-shape generation.
\newblock In \emph{CVPR}, pages 18603--18613, 2022.

\bibitem[Sanghi et~al.(2023{\natexlab{a}})Sanghi, Fu, Liu, Willis, Shayani, Khasahmadi, Sridhar, and Ritchie]{sanghi2023clip}
Aditya Sanghi, Rao Fu, Vivian Liu, Karl~DD Willis, Hooman Shayani, Amir~H Khasahmadi, Srinath Sridhar, and Daniel Ritchie.
\newblock Clip-sculptor: Zero-shot generation of high-fidelity and diverse shapes from natural language.
\newblock In \emph{Proceedings of the IEEE/CVF Conference on Computer Vision and Pattern Recognition}, pages 18339--18348, 2023{\natexlab{a}}.

\bibitem[Sanghi et~al.(2023{\natexlab{b}})Sanghi, Jayaraman, Rampini, Lambourne, Shayani, Atherton, and Taghanaki]{sanghi2023sketch}
Aditya Sanghi, Pradeep~Kumar Jayaraman, Arianna Rampini, Joseph Lambourne, Hooman Shayani, Evan Atherton, and Saeid~Asgari Taghanaki.
\newblock Sketch-a-shape: Zero-shot sketch-to-3d shape generation.
\newblock \emph{arXiv preprint arXiv:2307.03869}, 2023{\natexlab{b}}.

\bibitem[Schwarz et~al.(2022)Schwarz, Sauer, Niemeyer, Liao, and Geiger]{schwarz2022voxgraf}
Katja Schwarz, Axel Sauer, Michael Niemeyer, Yiyi Liao, and Andreas Geiger.
\newblock Voxgraf: Fast 3d-aware image synthesis with sparse voxel grids.
\newblock \emph{Advances in Neural Information Processing Systems}, 35:\penalty0 33999--34011, 2022.

\bibitem[Selvaraju et~al.(2021)Selvaraju, Nabail, Loizou, Maslioukova, Averkiou, Andreou, Chaudhuri, and Kalogerakis]{selvaraju2021buildingnet}
Pratheba Selvaraju, Mohamed Nabail, Marios Loizou, Maria Maslioukova, Melinos Averkiou, Andreas Andreou, Siddhartha Chaudhuri, and Evangelos Kalogerakis.
\newblock Buildingnet: Learning to label 3d buildings.
\newblock In \emph{Proceedings of the IEEE/CVF International Conference on Computer Vision}, pages 10397--10407, 2021.

\bibitem[Shi et~al.(2023)Shi, Wang, Ye, Long, Li, and Yang]{shi2023mvdream}
Yichun Shi, Peng Wang, Jianglong Ye, Mai Long, Kejie Li, and Xiao Yang.
\newblock Mvdream: Multi-view diffusion for 3d generation.
\newblock \emph{arXiv preprint arXiv:2308.16512}, 2023.

\bibitem[Shue et~al.(2023)Shue, Chan, Po, Ankner, Wu, and Wetzstein]{shue20233d}
J~Ryan Shue, Eric~Ryan Chan, Ryan Po, Zachary Ankner, Jiajun Wu, and Gordon Wetzstein.
\newblock {3D} neural field generation using triplane diffusion.
\newblock In \emph{CVPR}, pages 20875--20886, 2023.

\bibitem[Sohl-Dickstein et~al.(2015)Sohl-Dickstein, Weiss, Maheswaranathan, and Ganguli]{sohl2015deep}
Jascha Sohl-Dickstein, Eric Weiss, Niru Maheswaranathan, and Surya Ganguli.
\newblock Deep unsupervised learning using nonequilibrium thermodynamics.
\newblock In \emph{International conference on machine learning}, pages 2256--2265. PMLR, 2015.

\bibitem[Song and Ermon(2019)]{song2019generative}
Yang Song and Stefano Ermon.
\newblock Generative modeling by estimating gradients of the data distribution.
\newblock \emph{Advances in neural information processing systems}, 32, 2019.

\bibitem[Stojanov et~al.(2021)Stojanov, Thai, and Rehg]{stojanov2021using}
Stefan Stojanov, Anh Thai, and James~M Rehg.
\newblock Using shape to categorize: Low-shot learning with an explicit shape bias.
\newblock In \emph{Proceedings of the IEEE/CVF conference on computer vision and pattern recognition}, pages 1798--1808, 2021.

\bibitem[Su et~al.(2015)Su, Maji, Kalogerakis, and Learned-Miller]{su2015multi}
Hang Su, Subhransu Maji, Evangelos Kalogerakis, and Erik Learned-Miller.
\newblock Multi-view convolutional neural networks for 3d shape recognition.
\newblock In \emph{Proceedings of the IEEE international conference on computer vision}, pages 945--953, 2015.

\bibitem[Sun et~al.(2020)Sun, Wang, Liu, Siegel, and Sarma]{sun2020pointgrow}
Yongbin Sun, Yue Wang, Ziwei Liu, Joshua Siegel, and Sanjay Sarma.
\newblock Pointgrow: Autoregressively learned point cloud generation with self-attention.
\newblock In \emph{Proceedings of the IEEE/CVF Winter Conference on Applications of Computer Vision}, pages 61--70, 2020.

\bibitem[Verma et~al.(2018)Verma, Boyer, and Verbeek]{verma2018feastnet}
Nitika Verma, Edmond Boyer, and Jakob Verbeek.
\newblock Feastnet: Feature-steered graph convolutions for 3d shape analysis.
\newblock In \emph{Proceedings of the IEEE conference on computer vision and pattern recognition}, pages 2598--2606, 2018.

\bibitem[Vishwanath et~al.(2009)Vishwanath, Gupta, Vahdat, and Yocum]{vishwanath2009modelnet}
Kashi~Venkatesh Vishwanath, Diwaker Gupta, Amin Vahdat, and Ken Yocum.
\newblock Modelnet: Towards a datacenter emulation environment.
\newblock In \emph{2009 IEEE Ninth International Conference on Peer-to-Peer Computing}, pages 81--82. IEEE, 2009.

\bibitem[Wang et~al.(2019)Wang, Sun, Liu, Sarma, Bronstein, and Solomon]{wang2019dynamic}
Yue Wang, Yongbin Sun, Ziwei Liu, Sanjay~E Sarma, Michael~M Bronstein, and Justin~M Solomon.
\newblock Dynamic graph cnn for learning on point clouds.
\newblock \emph{ACM Transactions on Graphics (tog)}, 38\penalty0 (5):\penalty0 1--12, 2019.

\bibitem[Willis et~al.(2021)Willis, Pu, Luo, Chu, Du, Lambourne, Solar-Lezama, and Matusik]{willis2021fusion}
Karl~DD Willis, Yewen Pu, Jieliang Luo, Hang Chu, Tao Du, Joseph~G Lambourne, Armando Solar-Lezama, and Wojciech Matusik.
\newblock Fusion 360 gallery: A dataset and environment for programmatic cad construction from human design sequences.
\newblock \emph{ACM Transactions on Graphics (TOG)}, 40\penalty0 (4):\penalty0 1--24, 2021.

\bibitem[Wu et~al.(2016)Wu, Zhang, Xue, Freeman, and Tenenbaum]{wu2016learning}
Jiajun Wu, Chengkai Zhang, Tianfan Xue, Bill Freeman, and Josh Tenenbaum.
\newblock Learning a probabilistic latent space of object shapes via 3d generative-adversarial modeling.
\newblock \emph{Advances in neural information processing systems}, 29, 2016.

\bibitem[Wu et~al.(2021)Wu, Xiao, and Zheng]{wu2021deepcad}
Rundi Wu, Chang Xiao, and Changxi Zheng.
\newblock Deepcad: A deep generative network for computer-aided design models.
\newblock In \emph{Proceedings of the IEEE/CVF International Conference on Computer Vision}, pages 6772--6782, 2021.

\bibitem[Wu et~al.(2018)Wu, Wu, Gkioxari, and Tian]{wu2018building}
Yi Wu, Yuxin Wu, Georgia Gkioxari, and Yuandong Tian.
\newblock Building generalizable agents with a realistic and rich 3d environment.
\newblock \emph{arXiv preprint arXiv:1801.02209}, 2018.

\bibitem[Wu et~al.(2015)Wu, Song, Khosla, Yu, Zhang, Tang, and Xiao]{wu20153d}
Zhirong Wu, Shuran Song, Aditya Khosla, Fisher Yu, Linguang Zhang, Xiaoou Tang, and Jianxiong Xiao.
\newblock 3d shapenets: A deep representation for volumetric shapes.
\newblock In \emph{Proceedings of the IEEE conference on computer vision and pattern recognition}, pages 1912--1920, 2015.

\bibitem[Xu et~al.(2022)Xu, Jiang, Wang, Fan, Wang, and Wang]{xu2022neurallift}
Dejia Xu, Yifan Jiang, Peihao Wang, Zhiwen Fan, Yi Wang, and Zhangyang Wang.
\newblock Neurallift-360: Lifting an in-the-wild 2d photo to a 3d object with 360 views.
\newblock \emph{arXiv e-prints}, pages arXiv--2211, 2022.

\bibitem[Xu et~al.(2023{\natexlab{a}})Xu, Wang, Cheng, Cao, Shan, Qie, and Gao]{xu2023dream3d}
Jiale Xu, Xintao Wang, Weihao Cheng, Yan-Pei Cao, Ying Shan, Xiaohu Qie, and Shenghua Gao.
\newblock Dream3d: Zero-shot text-to-3d synthesis using 3d shape prior and text-to-image diffusion models.
\newblock In \emph{Proceedings of the IEEE/CVF Conference on Computer Vision and Pattern Recognition}, pages 20908--20918, 2023{\natexlab{a}}.

\bibitem[Xu et~al.(2023{\natexlab{b}})Xu, Tan, Luan, Bi, Wang, Li, Shi, Sunkavalli, Wetzstein, Xu, et~al.]{xu2023dmv3d}
Yinghao Xu, Hao Tan, Fujun Luan, Sai Bi, Peng Wang, Jiahao Li, Zifan Shi, Kalyan Sunkavalli, Gordon Wetzstein, Zexiang Xu, et~al.
\newblock Dmv3d: Denoising multi-view diffusion using 3d large reconstruction model.
\newblock \emph{arXiv preprint arXiv:2311.09217}, 2023{\natexlab{b}}.

\bibitem[Yan et~al.(2022)Yan, Lin, Mitra, Lischinski, Cohen-Or, and Huang]{yan2022shapeformer}
Xingguang Yan, Liqiang Lin, Niloy~J Mitra, Dani Lischinski, Daniel Cohen-Or, and Hui Huang.
\newblock Shapeformer: Transformer-based shape completion via sparse representation.
\newblock In \emph{Proceedings of the IEEE/CVF Conference on Computer Vision and Pattern Recognition}, pages 6239--6249, 2022.

\bibitem[Yang et~al.(2019)Yang, Huang, Hao, Liu, Belongie, and Hariharan]{yang2019pointflow}
Guandao Yang, Xun Huang, Zekun Hao, Ming-Yu Liu, Serge Belongie, and Bharath Hariharan.
\newblock Pointflow: 3d point cloud generation with continuous normalizing flows.
\newblock In \emph{Proceedings of the IEEE/CVF international conference on computer vision}, pages 4541--4550, 2019.

\bibitem[Yu et~al.(2022)Yu, Xu, Koh, Luong, Baid, Wang, Vasudevan, Ku, Yang, Ayan, et~al.]{yu2022scaling}
Jiahui Yu, Yuanzhong Xu, Jing~Yu Koh, Thang Luong, Gunjan Baid, Zirui Wang, Vijay Vasudevan, Alexander Ku, Yinfei Yang, Burcu~Karagol Ayan, et~al.
\newblock Scaling autoregressive models for content-rich text-to-image generation.
\newblock \emph{arXiv preprint arXiv:2206.10789}, 2\penalty0 (3):\penalty0 5, 2022.

\bibitem[Zeng et~al.(2022)Zeng, Vahdat, Williams, Gojcic, Litany, Fidler, and Kreis]{zeng2022lion}
Xiaohui Zeng, Arash Vahdat, Francis Williams, Zan Gojcic, Or Litany, Sanja Fidler, and Karsten Kreis.
\newblock Lion: Latent point diffusion models for 3d shape generation.
\newblock \emph{arXiv preprint arXiv:2210.06978}, 2022.

\bibitem[Zhang et~al.(2022)Zhang, Nie{\ss}ner, and Wonka]{zhang20223dilg}
Biao Zhang, Matthias Nie{\ss}ner, and Peter Wonka.
\newblock 3dilg: Irregular latent grids for 3d generative modeling.
\newblock \emph{Advances in Neural Information Processing Systems}, 35:\penalty0 21871--21885, 2022.

\bibitem[Zhang et~al.(2023{\natexlab{a}})Zhang, Tang, Niessner, and Wonka]{zhang20233dshape2vecset}
Biao Zhang, Jiapeng Tang, Matthias Niessner, and Peter Wonka.
\newblock 3dshape2vecset: A 3d shape representation for neural fields and generative diffusion models.
\newblock \emph{arXiv preprint arXiv:2301.11445}, 2023{\natexlab{a}}.

\bibitem[Zhang et~al.(2023{\natexlab{b}})Zhang, Tang, Niessner, and Wonka]{zhang2023vec}
Biao Zhang, Jiapeng Tang, Matthias Niessner, and Peter Wonka.
\newblock 3{DS}hape2{V}ec{S}et: A 3{D} shape representation for neural fields and generative diffusion models.
\newblock 42\penalty0 (4), 2023{\natexlab{b}}.

\bibitem[Zheng et~al.(2022)Zheng, Liu, Wang, and Tong]{zheng2022sdf}
Xinyang Zheng, Yang Liu, Pengshuai Wang, and Xin Tong.
\newblock Sdf-stylegan: Implicit sdf-based stylegan for 3d shape generation.
\newblock In \emph{Computer Graphics Forum}, pages 52--63. Wiley Online Library, 2022.

\bibitem[Zheng et~al.(2023)Zheng, Pan, Wang, Tong, Liu, and Shum]{zheng2023locally}
Xin-Yang Zheng, Hao Pan, Peng-Shuai Wang, Xin Tong, Yang Liu, and Heung-Yeung Shum.
\newblock Locally attentional sdf diffusion for controllable 3d shape generation.
\newblock \emph{arXiv preprint arXiv:2305.04461}, 2023.

\bibitem[Zhou et~al.(2021)Zhou, Du, and Wu]{zhou20213d}
Linqi Zhou, Yilun Du, and Jiajun Wu.
\newblock 3d shape generation and completion through point-voxel diffusion.
\newblock In \emph{Proceedings of the IEEE/CVF International Conference on Computer Vision}, pages 5826--5835, 2021.

\bibitem[Zhou and Jacobson(2016)]{zhou2016thingi10k}
Qingnan Zhou and Alec Jacobson.
\newblock Thingi10k: A dataset of 10,000 3d-printing models.
\newblock \emph{arXiv preprint arXiv:1605.04797}, 2016.

\bibitem[Zuffi et~al.(2017)Zuffi, Kanazawa, Jacobs, and Black]{zuffi2017smal}
Silvia Zuffi, Angjoo Kanazawa, David Jacobs, and Michael~J. Black.
\newblock {3D} menagerie: Modeling the {3D} shape and pose of animals.
\newblock In \emph{IEEE Conf. on Computer Vision and Pattern Recognition (CVPR)}, 2017.

\end{thebibliography}
}


\end{document}